\documentclass[preprint,12pt]{elsarticle}

\usepackage{amssymb}
\usepackage{amsmath}

\usepackage{xurl}
\usepackage{hyperref}

\usepackage[acronym]{glossaries}

\usepackage{graphicx}
\usepackage{subcaption}
\usepackage{rotating}

\usepackage{listings}
\usepackage{xcolor}
\lstdefinelanguage{yaml}{
  keywords={true, false, null, !!int, !!str},
  morekeywords={model_type, convs_params, denses_params, convs_dropout, denses_dropout, activation, use_batch_norm, epochs, batch_size, dataset, name, args, flat_features, random_seed},
  keywordstyle=\color{blue}\bfseries,
  commentstyle=\color{gray}\ttfamily,
  stringstyle=\color{red}\ttfamily,
  backgroundcolor=\color{lightgray!20},
  sensitive=false,
  comment=[l]{\#},
  morestring=[b]",
  morestring=[b]',
  literate={-}{{-}}1 {>}{{>}}1 {|}{{|}}1
           {:}{{:}}1 {,}{{,}}1
}
\usepackage{algorithm}
\usepackage{algpseudocode}

\usepackage{array}
\usepackage{makecell}
\usepackage{multirow}

\usepackage{pifont}
\newcommand{\cmark}{\ding{51}}

\glsdisablehyper
\newacronym{ml}{ML}{machine learning}
\newacronym{ai}{AI}{artificial intelligence}
\newacronym{dl}{DL}{deep learning}
\newacronym{mlp}{MLP}{multi-layer perceptron}
\newacronym{eai}{eAI}{embedded artificial intelligence}
\newacronym{tinyml}{TinyML}{tiny machine learning}
\newacronym{tf}{TF}{TensorFlow}
\newacronym{tflite}{TFLite}{TensorFlow Lite}
\newacronym{tflm}{TFLM}{TensorFlow Lite Micro}
\newacronym{svm}{SVM}{support vector machine}
\newacronym{fc}{FC}{fully connected}
\newacronym{cnn}{CNN}{convolutional neural network}
\newacronym{rnn}{RNN}{recurrent neural network}
\newacronym{xgboost}{XGBoost}{extreme gradient boosting}
\newacronym{lightgbm}{LightGBM}{light gradient boosting machine}
\newacronym{automl}{AutoML}{automated machine learning}
\newacronym{nas}{NAS}{neural architecture search}
\newacronym{ide}{IDE}{integrated development environment}
\newacronym{plc}{PLC}{programmable logic controller}
\newacronym{sdk}{SDK}{software development kit}
\newacronym{npu}{NPU}{neural processing unit}
\newacronym{cntk}{CNTK}{Microsoft Cognitive Toolkit}
\newacronym{wandb}{W\&B}{Weights \& Biases}
\newacronym{fpu}{FPU}{floating-point unit}
\newacronym{mac}{MAC}{multiply-accumulate}
\newacronym{usmp}{USMP}{unified static memory planner}
\newacronym{gcc}{GCC}{GNU Compiler Collection}
\newacronym{tvm}{TVM}{Tensor Virtual Machine}

\newcommand{\resultswidth}{0.75\textwidth}

\journal{Journal of Systems Architecture}

\begin{document}

\begin{frontmatter}



\title{EdgeMark: An Automation and Benchmarking System for Embedded Artificial Intelligence Tools}


\author[DTU]{Mohammad Amin Hasanpour\corref{cor}}
\ead{moam@dtu.dk}

\author[AU,Grundfos]{Mikkel Kirkegaard}
\ead{mik@ece.au.dk}

\author[DTU]{Xenofon Fafoutis}
\ead{xefa@dtu.dk}

\affiliation[DTU]{organization={Department of Applied Mathematics and Computer Science, \\Technical University of Denmark (DTU)}, 
            city={Kgs. Lyngby},
            postcode={2800}, 
            country={Denmark}}
\affiliation[AU]{organization={Department of Electrical and Computer Engineering, \\Aarhus University}, 
            city={Aarhus},
            postcode={8200}, 
            country={Denmark}}
\affiliation[Grundfos]{organization={Grundfos Holding A/S}, 
            city={Bjerringbro},
            postcode={8850}, 
            country={Denmark}}

\cortext[cor]{Corresponding author.}

\begin{abstract}

The integration of \gls{ai} into embedded devices, a paradigm known as \gls{eai} or \gls{tinyml}, is transforming industries by enabling intelligent data processing at the edge. However, the many tools available in this domain leave researchers and developers wondering which one is best suited to their needs. This paper provides a review of existing \gls{eai} tools, highlighting their features, trade-offs, and limitations. Additionally, we introduce EdgeMark, an open-source automation system designed to streamline the workflow for deploying and benchmarking \gls{ml} models on embedded platforms. EdgeMark simplifies model generation, optimization, conversion, and deployment while promoting modularity, reproducibility, and scalability. Experimental benchmarking results showcase the performance of widely used \gls{eai} tools, including \gls{tflm}, Edge Impulse, Ekkono, and Renesas eAI Translator, across a wide range of models, revealing insights into their relative strengths and weaknesses. The findings provide guidance for researchers and developers in selecting the most suitable tools for specific application requirements, while EdgeMark lowers the barriers to adoption of \gls{eai} technologies.

\end{abstract}




\end{frontmatter}



\section{Introduction}
\label{sec:introduction}

\Gls{ai} is revolutionizing how systems process, analyze, and act on data. While historically rooted in cloud computing, the need for faster, more secure, and localized \gls{ai} processing has driven the rise of \gls{eai}. Unlike conventional \gls{ai}, \gls{eai} processes data at the device level, transforming devices into standalone intelligent systems. This shift enables a wide range of applications, from smart home devices and wearables to industrial automation and healthcare \cite{Tsoukas2024}.

However, embedding \gls{ai} into resource-constrained devices introduces challenges, including processing power limitations, memory constraints, and energy efficiency requirements \cite{Immonen2022}. To address these challenges, a variety of tools have been developed to facilitate the deployment of \gls{ml} models on embedded devices. While these tools simplify the development process, they exhibit significant trade-offs in functionality, ease of use, and performance across different platforms. A concise review and benchmarking of these tools is necessary to guide developers in making informed decisions.

Beyond individual tools, deploying \gls{ml} models on embedded devices typically involves a multi-stage workflow comprising model generation, optimization, conversion, and deployment. Completing these steps, often manually, can be time-intensive and error-prone, particularly when scaling across multiple models or iterations. Automation systems designed to streamline this workflow have the potential to significantly reduce development time and improve consistency.

In this landscape, the contributions of this paper are threefold:
\begin{enumerate}
    \item \textbf{Review of \gls{eai} Tools:} We provide an overview of existing \gls{eai} frameworks, outlining their supported features, strengths, and limitations.
    \item \textbf{EdgeMark Automation System:} We introduce EdgeMark, an open-source framework designed to automate the deployment and benchmarking workflow, simplifying model generation, optimization, conversion, and deployment on embedded platforms. The codebase for EdgeMark is available here: \url{https://github.com/Black3rror/EdgeMark}.
    \item \textbf{Benchmarking \gls{eai} Tools:} Through extensive experiments, we evaluate the performance of widely used \gls{eai} tools across a wide array of models, highlighting trade-offs and differences they exhibit in various scenarios.
\end{enumerate}

Through the combination of these contributions, we aim to not only contextualize the current capabilities of \gls{eai} tools but also provide practical guidance and help for deploying ML models on embedded systems. By facilitating modularity, scalability, and reproducibility, EdgeMark represents an important step toward addressing the challenges of \gls{eai} deployment and paving the way for the next wave of intelligent applications at the edge.

The remainder of this paper is organized as follows. Section \ref{sec:related_work} provides an overview of related works. Section \ref{sec:eAI_tools} discusses the key features, functionality, and limitations of \gls{eai} tools. In Section \ref{sec:automation}, we describe the automation system developed to streamline the use of these tools, including its architecture and use cases. Section \ref{sec:benchmark} outlines the benchmarking methodology and presents experimental results comparing the performance of the tools across multiple parameters. Finally, Section \ref{sec:conclusion} concludes the paper by summarizing the contributions and proposing directions for future work.

\section{Related Work}
\label{sec:related_work}

While many \gls{tinyml} surveys are not specifically focused on \gls{eai} tools, they often include these tools as part of their review \cite{Tsoukas2024, Immonen2022, Njor2024, Sanchez2020, Ray2022, Saha2022, Capogrosso2024, Abadade2023}. These surveys typically provide a brief overview of available tools, covering many important ones but lacking completeness. Given the rapid evolution of the field, some information happens to be missing or becomes outdated.

MLPerf Tiny \cite{Banbury2020} is a benchmark suite for \gls{tinyml} models, designed to provide a standardized method for evaluating the performance of \gls{tinyml} models across different platforms. It includes four benchmarks in widely used \gls{tinyml} applications. Although it has played a significant role in improving \gls{tinyml} tools, we identify two main limitations in its use for comparing \gls{eai} tools. First, it includes only four models, all of which are considered large for many embedded devices. As a result, it cannot provide a comprehensive view of the tools' performance across a wide range of models, especially given that many industrial applications of \gls{eai} involve small models requiring less than 100 kB of memory. Second, MLPerf Tiny does not impose fixed hardware or configuration settings, which makes it challenging to compare the performance of different tools, as each may be evaluated under varying test conditions.

Several studies have compared popular \gls{eai} tools. For instance, \cite{Osman2022} examines the performance of \gls{tflm} and STM32Cube.AI on a few of models, concluding that STM32Cube.AI outperforms \gls{tflm} but is limited to STM32 devices, whereas \gls{tflm} offers broader flexibility. In \cite{Wulfert2024}, the creators of AIfES compare its performance against \gls{tflm}, showing that AIfES outperforms \gls{tflm}, particularly on \gls{fc} models. Similarly, \cite{MLonMCU} contrasts the interpreter and non-interpreter versions of \gls{tflm} with microTVM. This work also introduces a third backend for TVM that optimizes RAM usage through \gls{usmp} and runtime improvements. Although these studies provide valuable insights, they generally focus on comparing two tools within constrained test environments, falling short of offering a comprehensive evaluation.

Few works address the automation of benchmarking systems for \gls{tinyml}. In \cite{MLonMCU}, the authors implement automation to a certain extent, allowing them to conduct numerous comparisons within a short timeframe. \cite{Baciu2024} adopts this principle at its core but focuses more on traditional \gls{ml} models and tools. Nevertheless, the need for a comprehensive automation system capable of handling the entire workflow for deploying \gls{ml} models on embedded devices remains evident.

\section{eAI Tools}
\label{sec:eAI_tools}

\Gls{ai} models are commonly represented in high-level formats such as TensorFlow, PyTorch, or ONNX. While these formats facilitate model development, they are not always ideal for deployment on resource-constrained devices. This is because they are designed with rich interpreters in mind and include numerous features that embedded systems might not support efficiently. Therefore, \gls{ai} models must be converted to formats that are optimized for execution on the target device. Additionally, there may be opportunities for optimizing model execution specific to the hardware.

A variety of tools exist for deploying \gls{ai} models on embedded devices. These tools take advantage of their understanding of the target device's architecture to optimize model execution. Due to the diverse range of hardware available, different tools are suited for different devices. For instance, STM32Cube.AI is widely used for deploying models on STM32 devices, whereas the eAI Translator is tailored to match the specific hardware features of Renesas microcontrollers.

These deployment tools typically follow one of two approaches: direct execution or interpreter-based execution. Some tools incorporate the model execution directly into the firmware, whereas others convert the model into a binary format and use a separate interpreter to execute it. The interpreter approach tends to offer faster and more flexible development, enabling features such as shared memory management for multiple models and the ability to update models without altering the firmware. In contrast, direct execution can be more efficient in terms of both execution speed and memory usage.

Beyond model conversion, some tool providers offer comprehensive solutions that include data collection and labeling, model generation and testing, and data preprocessing and postprocessing. This broader scope makes the term ``toolchain" or ``platform" more appropriate for these offerings. In the following subsections, we will discuss some of the most popular tools for \gls{eai}.

\subsection{TensorFlow Lite}
\Gls{tflite}, recently renamed LiteRT, is a lightweight, fast, and optimized framework designed specifically for deploying \gls{dl} models on mobile and embedded devices. It converts models into FlatBuffers, a compact and efficient binary format that minimizes overhead. During conversion, \gls{tflite} can apply various model optimization techniques, such as quantization, pruning, and clustering. The resulting model can be executed on devices using the \gls{tflite} interpreter or used as input for further optimization or conversion to the target device's native language.

While \gls{tflite} primarily targets mobile devices and more capable embedded systems, \gls{tflm} \cite{David2020}, developed by Google in 2018 as an open-source project, is specifically tailored for microcontrollers and other resource-constrained devices with only a few kilobytes of memory. Although the heterogeneity of the devices was mentioned as a reason for having many tools, \gls{tflm} carries the ambition to unify them under a single framework by following a general kernel-based approach, whereby each vendor can optimize the kernels for their specific hardware. This capability has led to \gls{tflm}'s widespread acceptance among researchers and developers. For instance, CMSIS-NN provides optimized kernels for ARM Cortex-M processors and is integrated into \gls{tflm}. Additionally, \gls{tflm} includes optimized implementations for ARM Ethos-U processor series, Cadence Xtensa processor architecture, CEVA neural processing units, and ARC processors.

\Gls{tflm} processes a \gls{tflite} model and converts it into C++ source code that can be compiled with any C++17-compliant compiler. It operates without requiring an operating system or dynamic memory allocation, with the bare interpreter using less than 2 kB of memory \cite{David2020}. Although the original design of \gls{tflm} utilizes an interpreter to execute the model, there have been few attempts for direct model execution on devices. \cite{MLonMCU} demonstrates the success of one of these works in lowering the memory usage and significantly improving the setup time for the model, although the execution time remains the same.

\subsection{Edge Impulse}
Edge Impulse \cite{hymel2023edgeimpulsemlopsplatform}, founded in 2019, is a cloud-based platform that offers a comprehensive toolchain for developing, training, and deploying \gls{ml} models on embedded devices. It has evolved to support a wide array of devices and sensors, offering both free and enhanced professional and enterprise versions with additional features.

One of the key features of Edge Impulse is its flexibility with datasets. Having a dataset in their platform allows the users to find and train a model that fits their data best. Users can upload their own datasets or connect a device directly to the platform to collect data in real time. The platform simplifies data labeling and also supports data augmentation to expand the dataset and improve model generalization.

For data preprocessing, Edge Impulse offers a variety of signal processing blocks, paving the way for efficient model training. Users can choose from a range of \gls{ml} algorithms, including neural networks, logistic regression, \gls{svm}, random forest, \gls{xgboost}, and \gls{lightgbm}. The platform's EON Compiler stands out as an interpreter-less code generator for \gls{tflm}, reducing memory usage while maintaining kernel efficacy \cite{EdgeImpulseEON}. Recently, the EON Compiler has provided enterprise users with the option to trade off between memory usage and execution time. To achieve better performance, Edge Impulse supports 8-bit quantization.

Deployment flexibility is another strength of Edge Impulse. Users can output C++ source code for compilation on most embedded devices or generate a library for a specific \gls{ide} or a binary for direct device flashing. The platform supports an extensive range of hardware, from microcontrollers like the Arduino Nano 33 BLE and ESP32 to more powerful devices such as the Raspberry Pi and even Nvidia Jetson Nano. Additionally, Edge Impulse provides an estimation of the model's executation time, flash size, and RAM usage on the target device.

\subsection{microTVM}
microTVM \cite{MicroTVM} is a part of the Apache \gls{tvm} project, which is an open-source \gls{dl} compiler stack used to optimize and deploy \gls{ml} models on a variety of hardware platforms. microTVM is specifically designed to support deployment on microcontrollers and other resource-constrained devices.

microTVM uses the \gls{tvm} stack to compile \gls{ml} models into highly optimized code. The compilation involves several optimization steps such as operator fusion, memory planning, and hardware-specific optimizations. Binding with AutoTVM, a \gls{tvm} module for search and tuning the performance of operators on a specific hardware target, allows microTVM to generate highly optimized operators for the target device. For tuning tasks, AutoTVM typically requires some level of interaction with the target device, such as connecting to the microcontroller via USB-JTAG and accessing a cross-compiler and an on-chip debugger, to evaluate the performance of generated schedules.

To further enhance model efficiency, microTVM supports quantization. It does not require an operating system or dynamic memory allocation, making it suitable for bare-metal deployment. Although \gls{tflm} uses hand-optimized kernels like CMSIS-NN to ensure maximum efficacy and microTVM automatically generates optimized operators for the specific use case, microTVM is able to achieve comparable performance to \gls{tflm} \cite{Weber2020}.

\subsection{STM32Cube.AI}
STM32Cube.AI \cite{STM32CubeAI} is a powerful tool developed by STMicroelectronics in 2018 aimed at simplifying the deployment of neural network models on STM32 microcontrollers. While its source code is proprietary, the tool is freely accessible to STM32 developers. STM32Cube.AI integrates seamlessly with STM32CubeMX and can be accessed through graphical interface, command line interface, or online.

This tool efficiently converts models from popular frameworks such as TensorFlow, Keras, TFLite, PyTorch, Matlab, and Scikit-learn into C code. It employs techniques such as 8-bit or mixed-precision quantization, alongside graph and memory optimizations, to ensure optimal performance. In one of their comparative tests against \gls{tflm}, STM32Cube.AI has been shown to achieve, on average, 36\% faster execution time, 24\% smaller flash size, and 26\% reduced RAM usage for image classification and visual wake words models, as part of the MLPerf Tiny benchmark \cite{STM32CubeAISolution}.

Beyond its conversion capabilities, STM32Cube.AI offers users a comprehensive model zoo and supports them with numerous tutorials and examples. To aid user comprehension, it provides a visual representation of models and their layers, accompanied by detailed memory usage breakdown for each layer. Another interesting feature of STM32Cube.AI is the possibility of testing the models on real STM32 boards in their board farm. This will give users a measure of the model's latency, accuracy, and memory footprint on various STM32 devices, which can be helpful in choosing the right device for their application. Additionally, STM32Cube.AI supports storing model parameters and activations in external memory, which can be useful for models with large memory requirements.

\subsection{Renesas eAI Translator}
Renesas Electronics Corporation, a prominent Japanese semiconductor manufacturer, introduced the eAI Translator \cite{eAITranslator} in 2017. This tool allows for the deployment of \gls{ml} models on Renesas microcontrollers and operates as a plugin for Renesas' \gls{ide} for embedded systems development, e2 studio. The eAI Translator generates C code based on models provided by the user, supporting frameworks such as TensorFlow, Keras, TFLite, and PyTorch.

The tool is compatible with a wide range of microcontroller families, including RL78, RX, RA, RZ, and Renesas Synergy. It executes models directly on the microcontroller without requiring an interpreter. Additionally, eAI Translator supports 8-bit quantization with \gls{tflite} models and offers a specific version of the CMSIS-NN library, optimized for Renesas RX and RA architectures.

One notable feature of eAI Translator is its user-friendly interface, which simplifies the model conversion process. Users can decide whether to prioritize execution speed or RAM usage, a choice that determines how model parameters are allocated in either RAM or ROM. This flexibility allows users to execute small models even faster by storing them in RAM.

\subsection{Ekkono}
\label{subsec:ekkono}
Founded in 2016, Ekkono \cite{Ekkono} is a Swedish company that offers a platform for developing and deploying \gls{ml} models on embedded devices. A unique feature of Ekkono is its on-device learning capability, which sets it apart from many other tools. This feature allows models to adapt to their environment and improve their accuracy over time. While the computational and memory costs of this technique might raise some concerns, its potential to enable new applications at the edge is undeniable.

Unlike most tools that take in pretrained models in formats such as TensorFlow, ONNX, or TFLite, Ekkono requires users to develop and train models directly within its platform. This necessitates learning the Ekkono-specific model development process. Ekkono Primer is their main software library to develop models. Once models are developed, they can be deployed to edge devices using either Ekkono Edge or Ekkono Crystal.

Ekkono Edge is a C++ library designed for devices with a few megabytes of memory, such as communication gateways or \glspl{plc}. In contrast, Ekkono Crystal is a C library with limited functionality, optimized for microcontrollers that have only a few kilobytes of memory. Ekkono Crystal only relies on standard C libraries and can be compiled with any C99-compliant compiler. This library only supports linear regression and \gls{mlp} models with Sigmoid, Tanh, and LeakyRelu nonlinearities, targeting regression tasks.

Additionally, Ekkono offers Ekkono Spectral for signal processing needs. The platform also includes features like data preprocessing, \gls{automl}, sensitivity analysis, model change detection, conformal prediction, and anomaly detection. Ekkono relies on an interpreter to execute models on the edge. While Ekkono's platform is not free and requires licensing, it is well-documented, and users find it easy to use once they become familiar with its development process.

\subsection{ARM-NN}
Introduced by ARM in 2018, ARM-NN \cite{ArmNN} is an open-source project that targets more capable devices operating on Linux or Android. This \gls{sdk} is specifically designed to optimize performance on hardware architectures like ARM Cortex-A CPUs, ARM Mali GPUs, and ARM Ethos \glspl{npu}. It supports popular \gls{ml} frameworks, including TensorFlow, Caffe, and ONNX, leveraging the Compute Library for efficient hardware acceleration. Similar to CMSIS-NN, ARM-NN aims to enhance the deployment of neural networks across a range of ARM devices.

\subsection{Embedded Learning Library (ELL)}
ELL \cite{ELL}, an open-source project launched by Microsoft Research in 2017, aims to simplify the deployment of \gls{ml} models on resource-constrained platforms. It is particularly focused on small single-board computers such as the Raspberry Pi, and also supports devices like Arduino and micro:bit.

The project enables users to convert models from \gls{cntk}, Darknet, or ONNX into the ELL format, which serves as an intermediate representation. This format is then transformed into C++ code that can be compiled and executed on the target device. Alongside the C++ capabilities, ELL offers Python bindings for performing model inference.

It's important to note that ELL is currently inactive, and its codebase has recently been archived.

\subsection{NanoEdge AI Studio}
NanoEdge AI Studio \cite{NanoEdgeAIStudio}, an \gls{automl} platform developed by Cartesiam in 2020 and later acquired by STMicroelectronics, streamlines the process of finding optimal \gls{ml} models for user data. It accepts input data in text or CSV formats and supports various applications, including anomaly detection, outlier detection, classification, and regression tasks. Notably, the platform supports on-device learning for anomaly detection.

The output of the platform is a static library that can be linked to the user's C code and served to any ARM Cortex-M microcontroller. The generated models are lightweight, occupying under 20 kB of RAM and flash memory, and can execute in less than 20 milliseconds on a Cortex-M4 processor operating at 80 MHz \cite{NanoEdgeAIReq}. This platform features an intuitive and user-friendly interface and has recently become completely free \cite{NanoEdgeAINews}.

\subsection{uTensor}
uTensor \cite{uTensor} is a free and open-source project developed by Arm, designed to convert \gls{tf} models into C++ code. This code is primarily targeted at MBed OS, an open-source embedded operating system from Arm. In 2019, Arm announced that the uTensor project would be merged into \gls{tflm} \cite{Shelby2019}. More recently, Arm has also declared that MBed OS has reached its end of life, and it is no longer maintained or supported \cite{MbedUpdate}.

\subsection{Others}
We have identified approximately 50 tools related to \gls{eai}. In addition to the tools previously mentioned, we list the following: micromind \cite{MicromindToolkit}, ExecuTorch \cite{ExecuTorch}, Imagimob \cite{Imagimob}, OmniML \cite{OmniML}, EdgeML \cite{Jaiswal2023}, TinyEngine \cite{Lin2020}, FANN-on-MCU \cite{Wang2019}, AIfES \cite{FraunhoferAIfES}, TinyML gen \cite{Salerno2020}, CMix-NN \cite{Rusci2019}, Neurona \cite{Moretti2016}, hls4ml \cite{HLS4ML}, Silicon Labs MLTK \cite{MLTK}, Microsoft NNI \cite{Microsoft2021}, Core ML \cite{CoreML}, MNN \cite{Lv2022}, Glow \cite{Glow}, eIQ \cite{EIQ}, ONNX Runtime \cite{ONNXRuntime}, QKeras \cite{Coelho2020}, Larq \cite{Larq}, ONNC \cite{ONNC}, Latent AI \cite{LatentAI}, Plumerai \cite{Plumerai}, NNTool \cite{NNTool}, and DORY \cite{Burrello2021}. Additionally, as exemplified in \cite{Hasanpour2024}, traditional \gls{ml} algorithms hold significant promise for addressing many industrial applications. As a result, certain \gls{eai} tools have been specifically developed to support these algorithms. These include: sklearn-porter \cite{SklearnPorter}, weka-porter \cite{Morawiec2017}, m2cgen \cite{M2CGen}, MicroMLGen \cite{MicromlgenSklearn}, EmbML \cite{Tsutsui2022}, Qeexo \cite{QeexoAutoML}, and emlearn \cite{EmLearn}.

\subsection{Summary}

Major hardware vendors and prominent tech companies have each developed at least one tool to simplify the deployment of \gls{ml} models on embedded devices. Among the techniques employed in these tools, 8-bit integer quantization is the only one that is widely adopted. The main differentiating factor in performance among these tools is the use of hardware-specific optimizations. Beyond that, the size of the library may also be important, particularly for smaller models.

An ideal tool should give users maximum flexibility, allowing them to bring their own models from popular frameworks, quickly create models from scratch within the tool, or provide data for the tool to automatically discover the best models. \Gls{nas}, when designed to account for hardware performance metrics, is a valuable feature for identifying optimal models \cite{howard2019searching, cai2019once, gambella2022cnas, lomurno2024pomonag}. In addition to quantization, other optimization techniques, currently absent in most tools, should be considered. These include structured pruning for general use cases, unstructured pruning for accelerators that support it, neuron merging \cite{liu2021deep}, knowledge distillation \cite{hinton2015distilling}, and early exit networks \cite{ghanathe2022t}. The tool should also support a wide range of models and hardware platforms. On-device learning, supported by techniques specifically designed to minimize resource requirements \cite{song2022, pavan2024tybox, ren2021tinyol}, can play a key role in enabling \gls{eai} to expand into many new applications. Finally, the tool should be user-friendly, offering a simple and intuitive interface along with detailed documentation and tutorials.

\section{Automation of eAI Tools}
\label{sec:automation}

Deploying \gls{ml} models on embedded devices typically involves several steps, such as model generation, optimization, conversion, and deployment, which we refer to as function blocks. Each step can be time-consuming and rigorous, especially when managing multiple models or devices. This process is further complicated by the limited resources available on embedded devices, which demand careful consideration of the model's performance metrics and might necessitate multiple iterations of this workflow. To address these challenges, we propose EdgeMark, an open-source automation system that streamlines the deployment of \gls{ml} models on embedded devices.

\begin{figure*}[tbp]
  \centering
  \includegraphics{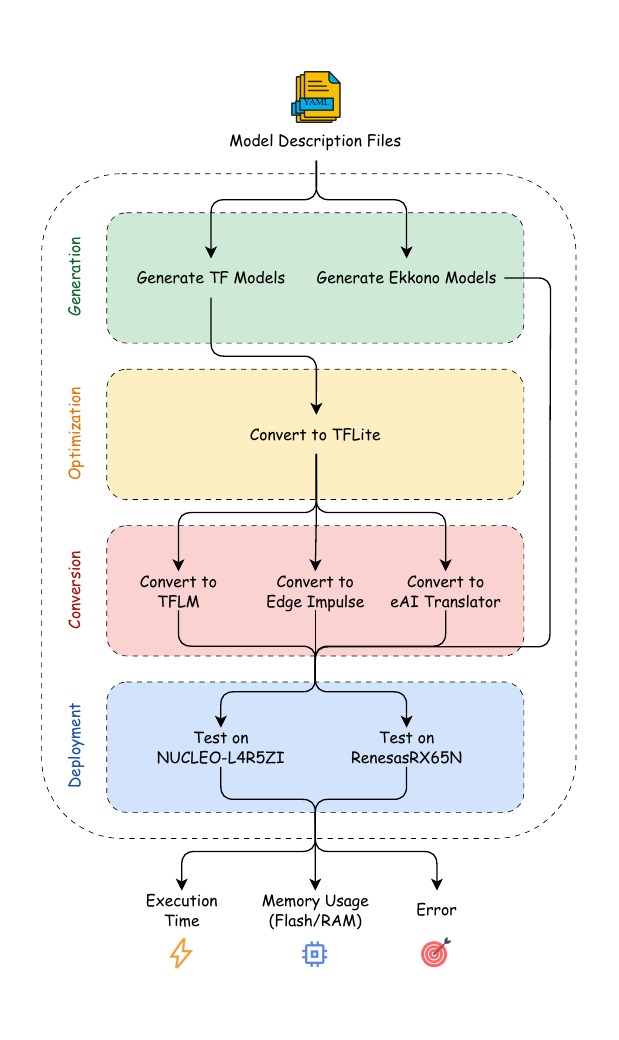}
  \caption{EdgeMark modules and their interconnections.}
  \label{fig:EdgeMark_modules}
\end{figure*}

EdgeMark follows a modular design that encompasses the entire pipeline of the aforementioned function blocks. Currently, EdgeMark includes modules whose correspondence to each function block is illustrated in Table~\ref{tab:modules_vs_function_blocks} and Fig.~\ref{fig:EdgeMark_modules}. These modules are:
\begin{itemize}
    \item \textbf{Generate TF Models:} Generates, compiles, trains, and evaluates TensorFlow models based on user-provided configuration files.
    \item \textbf{Generate Ekkono Models:} As discussed in Section \ref{subsec:ekkono}, Ekkono models require users to operate in a specific environment with a different workflow. This module simplifies the process by combining generation, training, evaluation, and conversion into a single step. It uses the same configuration files as the \textit{Generate TF Models} module and primarily produces Ekkono Edge and Ekkono Crystal models.
    \item \textbf{Convert to TFLite:} This module converts \gls{tf} models to \gls{tflite} format, offering various optimization options during the conversion process.
    \item \textbf{Convert to TFLM:} Converts \gls{tflite} models into the \gls{tflm} format, generating C++ code for deployment.
    \item \textbf{Convert to Edge Impulse:} This module converts \gls{tflite} models into the Edge Impulse format, also resulting in C++ code.
    \item \textbf{Convert to eAI Translator:} Prepares the necessary data files for integration with the eAI Translator's C project.
    \item \textbf{Test on NUCLEO-L4R5ZI:} This module enables testing of \gls{tflm}, Edge Impulse, and Ekkono models on the NUCLEO-L4R5ZI board. The respective project will be compiled and flashed to the board. The results will be collected, analyzed, and presented to the user.
    \item \textbf{Test on RenesasRX65N:} This module functions similar to \textit{Test on NUCLEO-L4R5ZI}, but is modified for the Renesas RX65N board.
\end{itemize}
The modules are interconnected, as shown in Figure \ref{fig:EdgeMark_modules}. This modular design offers several advantages. First, it allows users to select only the modules relevant to their specific application, making the system more flexible. For example, if a user is only interested in generating and converting models to TFLite, they can use the \textit{Generate TF Models} and \textit{Convert to TFLite} modules. Second, it makes the system easily extendable to support new models, tools, or devices. For instance, support for STM32Cube.AI can be added by introducing a new module in the \textit{Conversion} block.

\newcolumntype{C}[1]{>{\centering\arraybackslash}m{#1}}
\begin{table*}[htbp]
  \caption{Correspondence between EdgeMark modules and function blocks.}
  \begin{center}
    \begin{tabular}{c|C{3em}|C{3em}|C{3em}|C{3em}}
      & Gen & Opt & Conv & Deploy \\ \hline
      \textit{Generate TF Models} & \cmark & & & \\ \hline
      \textit{Generate Ekkono Models} & \cmark & & \cmark & \\ \hline
      \textit{Convert to TFLite} & & \cmark & * & \\ \hline
      \textit{Convert to TFLM} & & & \cmark & \\ \hline
      \textit{Convert to Edge Impulse} & & & \cmark & \\ \hline
      \textit{Convert to eAI Translator} & & & \cmark & \\ \hline
      \textit{Test on NUCLEO-L4R5ZI} & & & & \cmark \\ \hline
      \textit{Test on Renesas RX65N} & & & & \cmark \\
      \multicolumn{5}{l}{\makecell[l]{\rule{0pt}{7mm}* While this module converts the model into a format suitable \\ for devices like mobile phones, it cannot be directly used with \\ devices supported by EdgeMark.}}
    \end{tabular}
    \label{tab:modules_vs_function_blocks}
  \end{center}
\end{table*}

Running the main script will display the project's main menu where users can select the set of modules they wish to execute. EdgeMark runs the selected modules sequentially, storing their outputs for subsequent modules and for user reference. The primary output from the final layer of the workflow includes the execution time, flash size, RAM usage, and deployment error of the model on the target device. Additionally, the system generates a report summarizing the workflow results.

If generation of models is a part of the selected modules, users need to provide configuration files written in YAML format. These files detail the model architecture, dataset, and specific hyperparameters. Code Snippet~\ref{code:config_file} is an example configuration file. In this example, each element of \texttt{convs\_params} is a list of three integers that define the number of channels, kernel size, and stride, respectively. A zero in the number of channels indicates a max-pooling layer, and if the kernel size is also zero, it represents a global average pooling layer. The \texttt{denses\_params} specifies the number of neurons in each dense layer. The model's output neurons and the loss function will be automatically set based on the dataset.

\newfloat{code}{htbp}{lop}
\floatname{code}{Code Snippet}

\floatstyle{plain}
\restylefloat{code}

\begin{code}[htb]
\centering
\begin{lstlisting}
model_type: "CNN"
convs_params: [
    [8, 3, 1],
    [0, 2, 2],
    [16, 3, 1],
    [0, 0, 0]
]
denses_params: [64, 16]
convs_dropout: 0.25
denses_dropout: 0.10
activation: "relu"
use_batch_norm: False
epochs: 50
batch_size: 32
dataset:
    name: "mnist"
    args:
        flat_features: False
random_seed: 42
\end{lstlisting}
\caption{Example of a user-defined configuration file for generating a CNN model.}
\label{code:config_file}
\end{code}

EdgeMark currently supports various models, such as \gls{fc}, \gls{cnn}, and \gls{rnn}, along with specialized models like MLPerf Tiny models. The system includes datasets like MNIST, CIFAR-10, ImageNet-V2, Visual Wake Words, Boston Housing, and more. In addition to \gls{tflite} conversion, users can choose among \gls{tflm}, Edge Impulse, Ekkono, and Renesas eAI Translator for converting models to source code. The models can be tested on NUCLEO-L4R5ZI and Renesas RX65N boards. The complete list of supported models, datasets, tools, and hardware is presented in Table~\ref{tab:EdgeMark_support_range}. It is important to note that, EdgeMark is designed for easy extensibility, allowing new models, datasets, tools, and devices to be incorporated with minimal effort.

\begin{table}[htbp]
  \caption{EdgeMark support range for models, datasets, tools, and hardware.}
  \begin{center}
    \begin{tabular}{|c|c|c|}
      \hline
      \multirow{10}{*}{\textbf{Models}} & \multicolumn{2}{c|}{FC} \\ \cline{2-3}
      & \multicolumn{2}{c|}{CNN} \\ \cline{2-3}
      & \multirow{3}{*}{RNN} & SimpleRNN \\ \cline{3-3}
      & & LSTM \\ \cline{3-3}
      & & GRU \\ \cline{2-3}
      & \multirow{4}{*}{MLPerf Tiny} & AE \\ \cline{3-3}
      & & DS CNN \\ \cline{3-3}
      & & ResNet \\ \cline{3-3}
      & & MBNet \\ \cline{2-3}
      & \multicolumn{2}{c|}{MBNetV2} \\ \hline \hline

      \multirow{11}{*}{\textbf{Datasets}} & \multicolumn{2}{c|}{MNIST} \\ \cline{2-3}
      & \multicolumn{2}{c|}{CIFAR-10} \\ \cline{2-3}
      & \multicolumn{2}{c|}{ImageNet-V2} \\ \cline{2-3}
      & \multicolumn{2}{c|}{Boston Housing} \\ \cline{2-3}
      & \multicolumn{2}{c|}{Visual Wake Words} \\ \cline{2-3}
      & \multicolumn{2}{c|}{Shakespeare} \\ \cline{2-3}
      & \multicolumn{2}{c|}{Sinus} \\ \cline{2-3}
      & \multicolumn{2}{c|}{ToyADMOS} \\ \cline{2-3}
      & \multirow{3}{*}{RandomSet} & Classification \\ \cline{3-3}
      & & Regression \\ \cline{3-3}
      & & Sequence \\ \hline \hline

      \multirow{4}{*}{\textbf{Tools}} & \multicolumn{2}{c|}{TFLM} \\ \cline{2-3}
      & \multicolumn{2}{c|}{Edge Impulse} \\ \cline{2-3}
      & \multicolumn{2}{c|}{Ekkono} \\ \cline{2-3}
      & \multicolumn{2}{c|}{Renesas eAI Translator} \\ \hline \hline

      \multirow{2}{*}{\textbf{Hardware}} & STM & NUCLEO-L4R5ZI \\ \cline{2-3}
      & Renesas & RX65N \\ \hline
    \end{tabular}
    \label{tab:EdgeMark_support_range}
  \end{center}
\end{table}

EdgeMark supports \gls{tflm} due to its popularity in deploying models on microcontrollers. Similarly, Edge Impulse has been integrated for its extensive features, flexibility, and user-friendly interface. Ekkono is also included in our suite for its lightweight library and unique on-device learning capability, which sets it apart in the field. To ensure a comprehensive comparison of performance, we have included Renesas eAI Translator. This allows us to evaluate general tools like \gls{tflm}, which may have a bias toward ARM Cortex-M processors, against more vendor-specific solutions.

In the remainder of this section, we delve deeper into the functionality of each module. \textit{Generate TF Models} generates a TensorFlow model for each configuration file available in a specific directory. This process results in the storage of TensorFlow models, their corresponding best and last weights, and training and evaluation logs, which can be reviewed using TensorBoard and \gls{wandb}, along with some supplementary files. Notably, in this module, users can select random dataset generators to create datasets with their desired characteristics. These generators allow users to complete the system's workflow and enable evaluation of performance metrics like execution time, memory usage, and deployment error, regardless of model correctness or accuracy.

The \textit{Convert to TFLite} module is responsible for transforming TensorFlow models into \gls{tflite} models. Beyond standard conversion, this module provides a range of optimization options, including six types of quantization (such as 8-bit integer and dynamic range quantization), pruning, clustering, and a collaborative optimization strategy that combines multiple optimizations.

\textit{Convert to TFLM/Edge Impulse/eAI Translator} converts \gls{tflite} models into their platform's specific C/C++ code. To minimize memory consumption, \gls{tflm} requires users to include only the operators necessary for running the model. The \textit{Convert to TFLM} module automates this process by identifying the operators used in the model and including them in the generated code.

\Gls{tflm} employs a static memory allocation strategy, which means users must define an arena size--a fixed memory buffer where all allocations occur during model execution. Determining the minimum viable arena size involves two steps: first, estimating the memory usage of the model, and second, running a search algorithm to refine the value. The goal is to find the smallest arena size that allows the model to run without encountering memory errors.

The initial memory estimation is based on the maximum memory demand of consecutive model layers. Specifically, it considers the sum of the output sizes of two consecutive layers that must coexist in memory, assuming no branching in the model architecture. This estimation provides a starting point for the search process.

To accurately determine the minimum arena size, we implement a search algorithm, a simplified version of which is outlined in Algorithm \ref{alg:min_arena_size}. The algorithm refines the arena size in increments determined by a resolution of 0.5, 1, or 2 kB, depending on the estimated value. While the initial estimate is expected to be close to the actual minimum in most cases, significant deviations can occur in certain scenarios. During experiments, the growth and shrinking parameters of the algorithm can be fine-tuned to accelerate convergence and improve efficiency in finding the optimal arena size.

\begin{algorithm*}
\caption{Find Minimum Arena Size}
\label{alg:min_arena_size}
\begin{algorithmic}[1]
  \State growth\_rate $\gets$ 1.25
  \State shrink\_rate $\gets$ 0.8
  \State growth\_steps $\gets$ 4
  \State shrink\_steps $\gets$ 4
  \State \textbf{Initialize} lower\_bound, upper\_bound, best\_guess $\gets$ \textbf{None}

  \Statex

  \Function{UpdateBounds}{guess}
    \State try guess on embedded device
    \If{guess is too low}
      \State lower\_bound $\gets$ guess
    \ElsIf{guess is too high}
      \State upper\_bound $\gets$ guess
    \ElsIf{guess is correct}
      \State best\_guess $\gets$ guess
      \State upper\_bound $\gets$ guess
    \EndIf
  \EndFunction

  \Statex

  \State \Call{UpdateBounds}{estimated\_arena\_size}

  \Statex

  \While{upper\_bound - lower\_bound $\neq$ 1}
    \If{lower\_bound exists and not upper\_bound}
      \State next\_guess $\gets$ max(lower\_bound $\times$ growth\_rate, lower\_bound $+$ growth\_steps)
    \ElsIf{upper\_bound exists and not lower\_bound}
      \State next\_guess $\gets$ min(upper\_bound $\times$ shrink\_rate, upper\_bound $-$ shrink\_steps)
    \Else
      \State next\_guess $\gets$ (lower\_bound + upper\_bound) / 2
    \EndIf

    \Statex

    \State \Call{UpdateBounds}{next\_guess}
  \EndWhile
\end{algorithmic}
\end{algorithm*}

For convenience, the \textit{Generate Ekkono Models} module accepts the same configuration files as the \textit{Generate TF Models} module. However, it is important to note that Ekkono supports only a limited subset of TensorFlow models. Specifically, it supports \gls{fc} models for regression tasks, excluding features like dropout or batch normalization. Additionally, Ekkono does not support optimization techniques such as quantization, pruning, or clustering. Once the models are generated, they are trained and evaluated using the Ekkono-specific workflow, with the resulting Ekkono Edge and Ekkono Crystal models stored as output.

To deploy these models on target devices, generic C/C++ projects are prepared for each platform and hardware. The generated code for models in previous modules adheres to the structure of these general projects; for example, they define the model's input and output size and type. Thus, for \textit{Test on NUCLEO-L4R5ZI} and \textit{Test on RenesasRX65N}, changing the model is as simple as replacing the model's source code with the new one. Afterwards, they compile and flash the project to the board, where the model is executed, and the results are collected. These results are then analyzed and presented to the user in a report, which includes the following key metrics:
\begin{itemize}
  \item \textbf{Execution time:} This metric measures how long it takes for the model to run on the target device. It is measured by iteratively running the model 10 times and taking the average. The same input was used across all iterations. In our testing environment, changing the input, having a warm-up phase, or re-running the test have negligible to no effect on the results. The standard deviation of the execution time was found to be close to zero, reflecting the consistency of the measurement.
  \item \textbf{Flash size:} This represents the amount of flash memory required by the model. The total flash memory usage is reported by the compiler, and the flash size for the model and its dependencies is obtained by subtracting the flash memory usage of a base project (a similar project without the model and its dependencies) from that of the project including the model.
  \item \textbf{RAM usage:} Similar to flash size, RAM usage quantifies the memory consumed by the model during execution. It is also based on the compiler's report, since all tools supported by EdgeMark follow a static memory allocation strategy.
  \item \textbf{Deployment error:} This is the error rate of the model when deployed on the target device. The deployment error is computed as the average normalized absolute difference between the model's output on the target device and its expected output generated by running the model on a PC. The model on PC can be either the basic version or optimized version (e.g., quantized version). We assume optimization as a part of conversion and always use the basic version of the model on PC. To calculate deployment error, we provide the model with 10 different inputs and report the average error rate.
\end{itemize}

\section{Benchmark of eAI Tools}
\label{sec:benchmark}

To evaluate the performance of the \gls{eai} tools, we conducted a series of experiments using EdgeMark. Additionally, we analyzed the impact of various factors, such as different quantization schemes and the presence of a \gls{fpu}. In total, we have performed thousands of tests, with each test taking an average of two to three minutes to complete, the majority of which was spent on cross-compilation\footnote{The cross-compilation was done on a machine with an Intel Core i7-11850H CPU.} and flashing the project to the boards.

EdgeMark employs several tools and platforms to accomplish its objective. Table~\ref{tab:tools_versions} summarizes their versions and purposes as used in the benchmark. Among these, STM32CubeIDE, STM32CubeCLT, Renesas e2 studio, and Renesas Flash Programmer, must be installed manually if the relevant modules are used. Additionally, an Edge Impulse account is necessary to utilize its corresponding module.

\begin{table*}[htbp]
  \caption{Tools and Platforms Used in EdgeMark}
  \begin{center}
    \begin{tabular}{|l|l|l|}
      \hline
      \textbf{Name}& \textbf{Version} & \textbf{Purpose} \\
      \hline
      TensorFlow & 2.15.0 & Model generation \\
      \hline
      TFLite & \makecell[l]{Coupled with \\ TensorFlow} & \makecell[l]{Model optimization and \\ conversion} \\
      \hline
      TFLM & Commit 42f4bb8 & Model conversion \\
      \hline
      Edge Impulse & \makecell[l]{1.56.13 \\ date: 12-09-2024} & Model conversion \\
      \hline
      Ekkono & 23.10 & Model generation and conversion \\
      \hline
      \makecell[l]{Renesas eAI \\ Translator} & 3.2.0 & Model conversion \\
      \hline
      STM32CubeIDE & 1.14.1 & Project setup and compilation \\
      \hline
      STM32CubeCLT & 2.16.0 & Flashing projects to the board \\
      \hline
      Renesas e2 studio & 24.1.1 & Project setup and compilation \\
      \hline
      \makecell[l]{Renesas \\ Flash Programmer} & 3.15.00 & Flashing projects to the board \\
      \hline
    \end{tabular}
    \label{tab:tools_versions}
  \end{center}
\end{table*}

For both \glspl{ide}, STM32CubeIDE and Renesas e2 studio, the projects were created using default settings, with the exception of the compiler optimizations, which were set to their maximum level. To execute \gls{tflm} models on the NUCLEO-L4R5ZI board, we integrated the CMSIS-NN library, which significantly accelerates the performance of 8-bit and 16-bit integer quantized models by leveraging the SIMD instructions of the ARM Cortex-M4 processor. Renesas has also provided a CMSIS-NN library optimized for their RX microcontrollers, which we have used to execute eAI Translator models on the Renesas RX65N board.

\subsection{Models}
\label{sec:models}
\begin{figure*}[tbp]
  \centering
  \begin{subfigure}[b]{\resultswidth}
    \includegraphics[width=\textwidth]{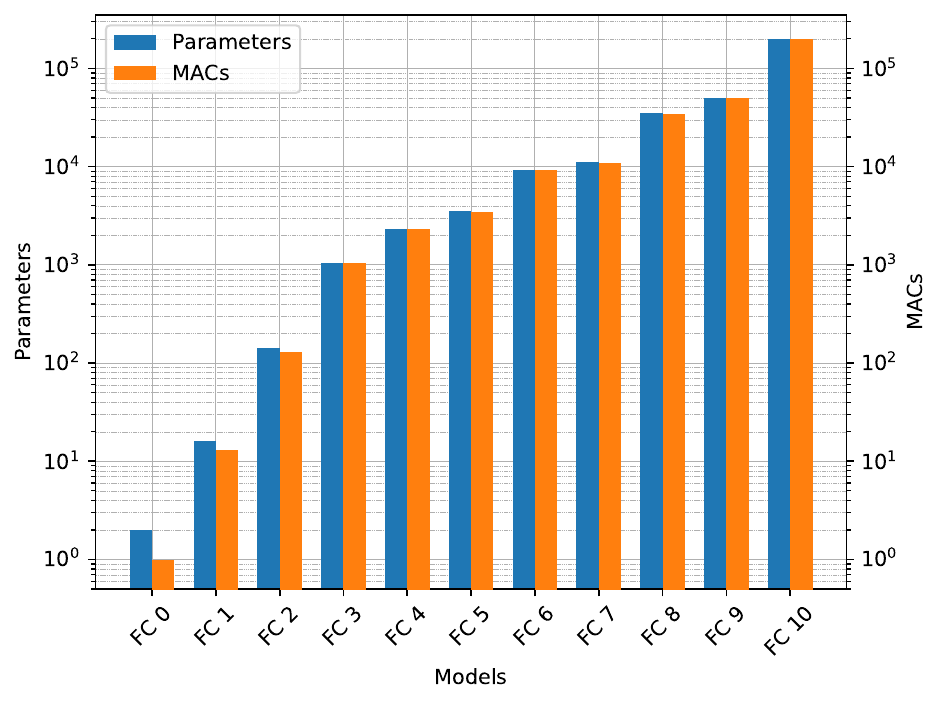}
    \label{fig:FC_stats}
  \end{subfigure}
  \begin{subfigure}[b]{\resultswidth}
    \vskip 5mm  
    \includegraphics[width=\textwidth]{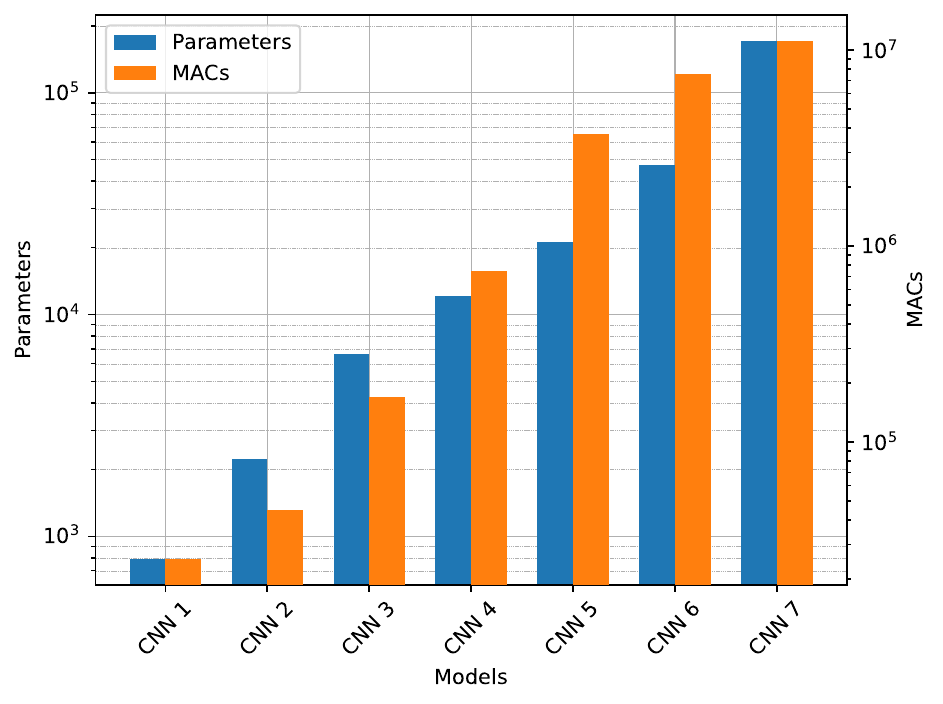}
    \label{fig:CNN_stats}
  \end{subfigure}
  \caption{Number of parameters and MACs for \gls{fc} and \gls{cnn} models.}
  \label{fig:FC_CNN_stats}
\end{figure*}

We selected a wide range of models to ensure a comprehensive evaluation of the tools' capabilities. These include 11 \gls{fc} models, 7 \gls{cnn} models, 7 \gls{rnn} models, and 4 models taken from the MLPerf Tiny benchmark. As illustrated in Fig.~\ref{fig:FC_CNN_stats}, the \gls{fc} and \gls{cnn} models are well distributed across a range of parameters and \glspl{mac}, ensuring that our evaluation covers both lightweight and computationally intensive models.

\textit{FC 0} serves as a minimal baseline, with an input size of 1, an output size of 1, and no hidden layers. It can be useful for analyzing the overhead of the library executing the model. Similarly, in the \gls{rnn} category, \textit{Simple 0}, which has the minimal structure of an \gls{rnn}, plays an analogous role. However, because \gls{rnn} models involve more complex operations, \textit{Simple 0} is much more demanding compared to \textit{FC 0}.

The \gls{rnn} models include five SimpleRNNs, one LSTM, and one GRU. All \gls{rnn} models process sequence lengths of 100, except for \textit{Simple 0}, which operates on a sequence length of 2. Two of the SimpleRNN-based models are \textit{Shakespeare 1} and \textit{Shakespeare 2} models that have an embedding layer before their \gls{rnn} units and are trained on the \texttt{tiny\_shakespeare} dataset \cite{Karpathy2015}. Detailed information about the \gls{rnn} models is presented in Table~\ref{tab:RNN_stats}.

\begin{table*}[htbp]
  \caption{Details of \gls{rnn} Models}
  \begin{center}
    \begin{tabular}{|l|l|l|r|r|}
      \hline
      \textbf{Model} & \textbf{RNN Units} & \textbf{RNN Input} & \textbf{Params} & \textbf{MACs} \\
      \hline
      Simple 0 & 1 & 1 & 5 & 9 \\
      \hline
      Simple 1 & 64 & 32 & 8,288 & 827,200 \\
      \hline
      Simple 2 & 128 & 64 & 32,960 & 3,292,800 \\
      \hline
      Shakespeare 1 & 64 & 32 & 12,513 & 1,056,300 \\
      \hline
      Shakespeare 2 & 128 & 64 & 37,249 & 3,321,900 \\
      \hline
      LSTM & 64 & 32 & 26,912 & 2,702,400 \\
      \hline
      GRU & 64 & 32 & 20,896 & 2,094,400 \\
      \hline
    \end{tabular}
    \label{tab:RNN_stats}
  \end{center}
\end{table*}

Lastly, Table~\ref{tab:MLPerf_stats} provides an overview of the MLPerf Tiny models used in this evaluation. By including this wide range of models, we aim to provide meaningful insights into the performance and efficiency of the tools across different neural network architectures.

\begin{table*}[htbp]
  \caption{Details of MLPerf Tiny Models}
  \begin{center}
    \begin{tabular}{|l|r|r|l|}
      \hline
      \textbf{Model} & \textbf{Params} & \textbf{MACs} & \textbf{Application} \\
      \hline
      AE (Autoencoder) & 269,992 & 269,156 & Anomaly detection \\
      \hline
      \makecell[l]{DS CNN (Depthwise \\ Separable CNN)} & 24,908 & 2,696,804 & Keyword spotting \\
      \hline
      ResNet & 78,666 & 12,561,054 & Image classification \\
      \hline
      MobileNet & 221,794 & 7,606,598 & Image classification \\
      \hline
    \end{tabular}
    \label{tab:MLPerf_stats}
  \end{center}
\end{table*}

\subsection{Experiments}
We have structured our analysis into ten experiments, five of which are described in this section, with the remaining five detailed in \ref{sec:appendix}. It is worth emphasizing that each experiment is designed to compare the general behavior of the test subjects, though there may be some exceptional cases that deviate from the trends. For a comprehensive overview of the results, readers are encouraged to refer to the project's documentation\footnote{\href{https://black3rror.github.io/EdgeMark}{https://black3rror.github.io/EdgeMark}}.

In general, each experiment focuses on the following discussion points: \textit{Model correctness}, which examines whether any models failed to execute or produced unacceptable deployment errors; \textit{execution time}, \textit{flash size}, and \textit{RAM usage}, which provide comparative insights into the respective metrics across study elements; and, finally, a \textit{conclusion} that summarizes the overall performance and behavior of the subjects under consideration.

\subsubsection{Quantizations}
In general, the following quantization schemes are available in EdgeMark:
\begin{itemize}
  \item \textbf{Dynamic:} Dynamic range quantization. Weights are quantized to 8-bit integers, while activations are stored in 32-bit floats. To accelerate inference, activations can be dynamically quantized to 8-bit integers and then dequantized back to 32-bit floats for storage \cite{GoogleDynamicQuant}.
  \item \textbf{Int8:} Full integer quantization. Both weights and activations are quantized to 8-bit integers. However, the input and output remain in 32-bit floats \cite{GoogleIntegerQuant}.
  \item \textbf{Int8 only:} Similar to \textit{int8}, but in this case, everything is quantized to integers without fallback to floats \cite{GoogleIntegerQuant}.
  \item \textbf{16x8:} To improve the accuracy of the quantized model, the activations are quantized to 16-bit integers while the weights are quantized to 8-bit integers. Similar to \textit{int8}, the input and output are left in 32-bit floats \cite{GoogleInt16Quant}.
  \item \textbf{16x8 int only:} Same as \textit{16x8}, but everything is quantized to integers without fallback to floats \cite{GoogleInt16Quant}.
  \item \textbf{Float 16:} Weights are quantized to 16-bit floats, reducing model size with minimal impact on accuracy \cite{GoogleFloat16Quant}.
\end{itemize}

In this study, we evaluate the performance of several quantization schemes using the \gls{tflm} platform. \Gls{tflm} was chosen because it is the only framework that supports the majority of available quantization methods. In particular, it supports all the abovementioned quantizations, except \textit{float 16}. This is while, Edge Impulse and eAI Translator only support the \textit{int8 only} quantization\footnote{The \textit{int8} quantization may also be supported, but it was not tested.}.

\begin{figure*}[tbp]
  \centering
  \begin{subfigure}[b]{\resultswidth}
    \includegraphics[width=\textwidth]{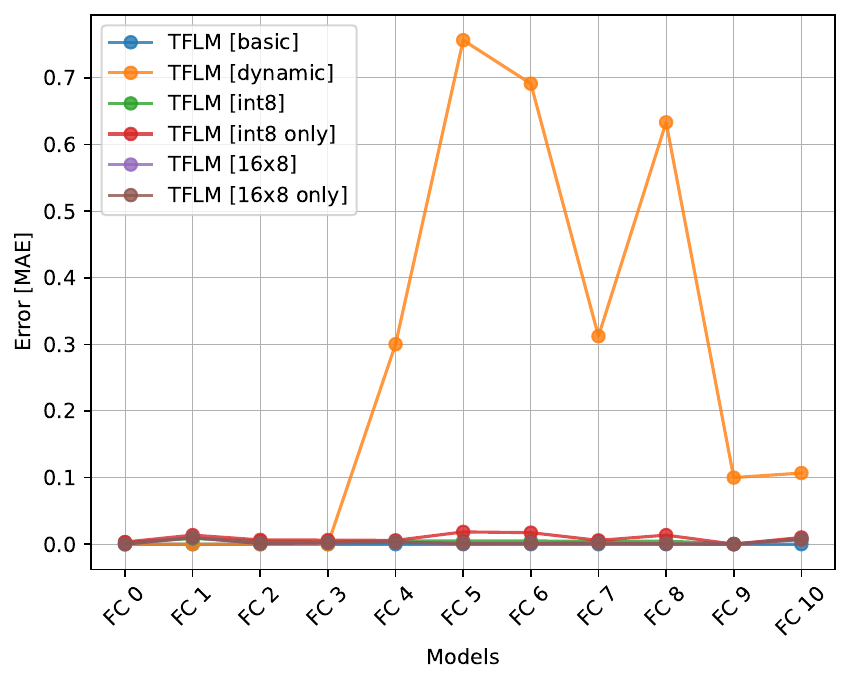}
    \caption{\Gls{fc} models.}
    \label{fig:study_quants_fc_error}
  \end{subfigure}
  \begin{subfigure}[b]{\resultswidth}
    \vskip 5mm  
    \includegraphics[width=\textwidth]{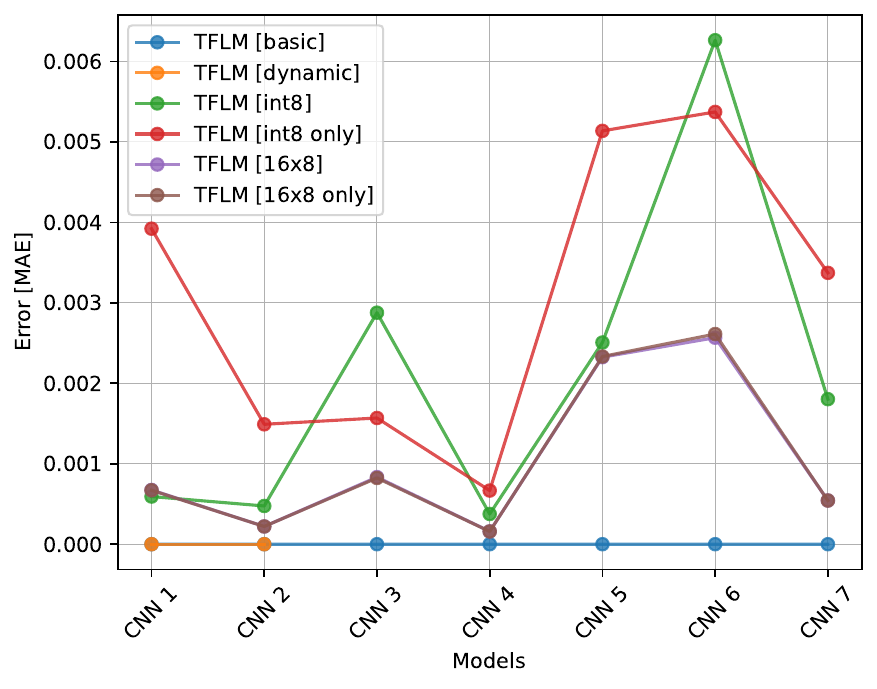}
    \caption{\Gls{cnn} models.}
    \label{fig:study_quants_cnn_error}
  \end{subfigure}
  \caption{Deployment error across various quantization schemes. The models were tested on NUCLEO-L4R5ZI.}
  \label{fig:study_quants_error}
\end{figure*}

\begin{figure*}[tbp]
  \centering
  \begin{subfigure}[b]{\resultswidth}
    \includegraphics[width=\textwidth]{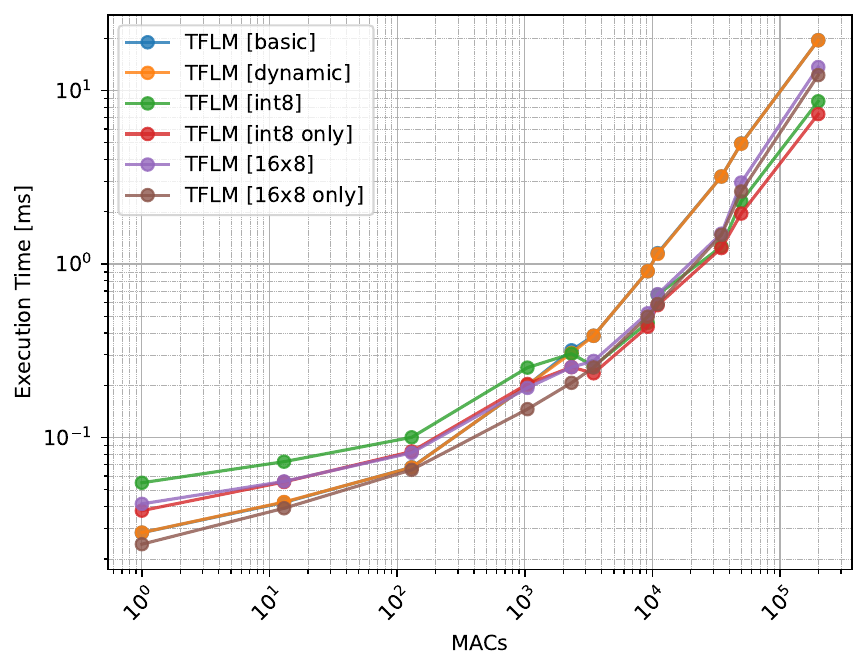}
    \caption{\Gls{fc} models.}
    \label{fig:study_quants_fc_exe}
  \end{subfigure}
  \begin{subfigure}[b]{\resultswidth}
    \vskip 5mm  
    \includegraphics[width=\textwidth]{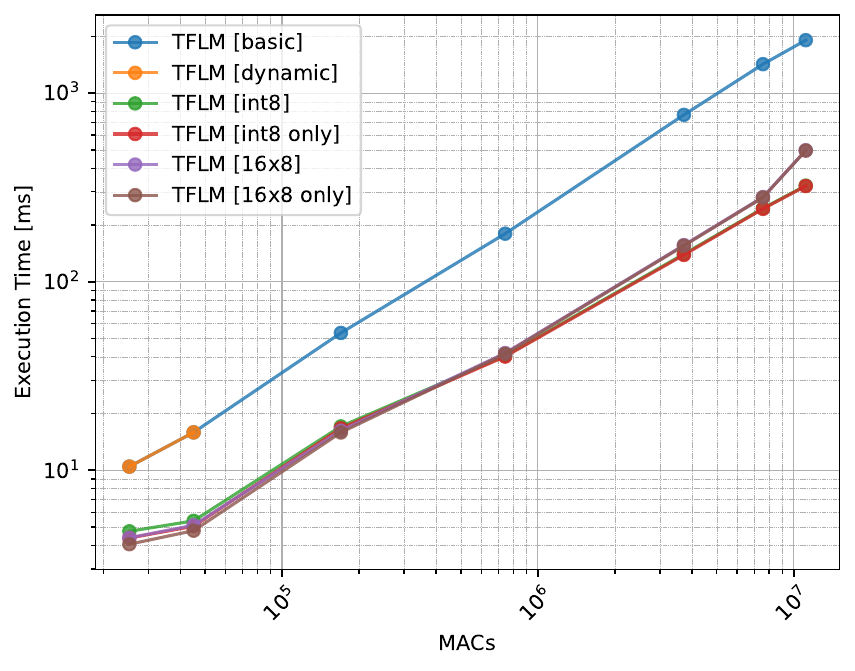}
    \caption{\Gls{cnn} models.}
    \label{fig:study_quants_cnn_exe}
  \end{subfigure}
  \caption{Average execution time across various quantization schemes. The models were tested on NUCLEO-L4R5ZI.}
  \label{fig:study_quants_exe}
\end{figure*}

\begin{figure*}[tbp]
  \centering
  \begin{subfigure}[b]{\resultswidth}
    \includegraphics[width=\textwidth]{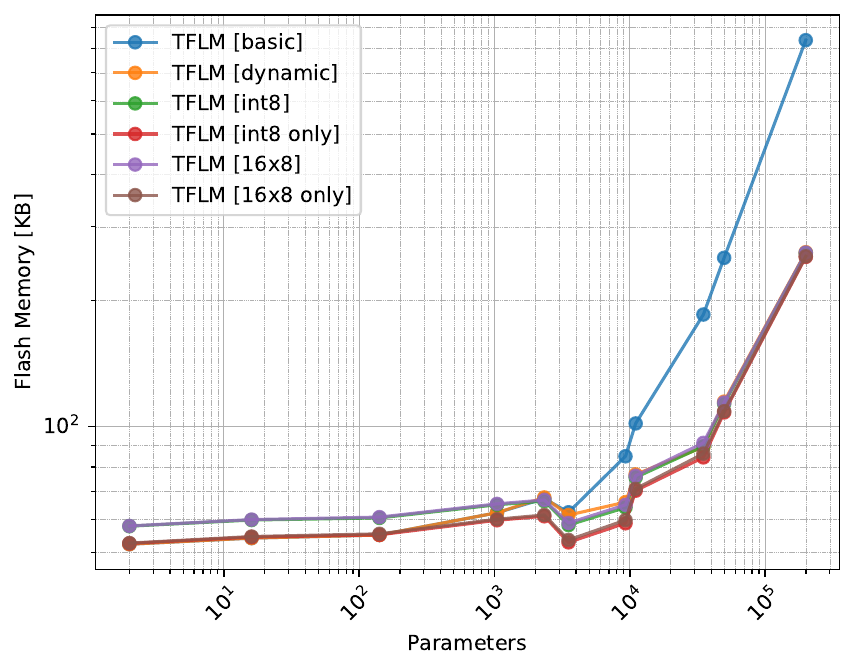}
    \caption{\Gls{fc} models.}
    \label{fig:study_quants_fc_flash}
  \end{subfigure}
  \begin{subfigure}[b]{\resultswidth}
    \vskip 5mm  
    \includegraphics[width=\textwidth]{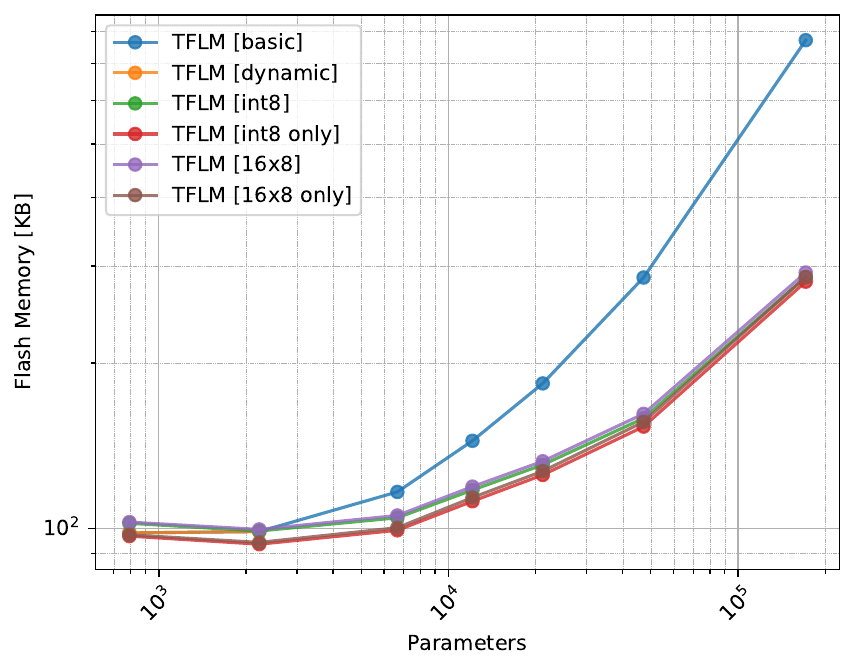}
    \caption{\Gls{cnn} models.}
    \label{fig:study_quants_cnn_flash}
  \end{subfigure}
  \caption{Flash memory usage for various quantization schemes. The models were tested on NUCLEO-L4R5ZI.}
  \label{fig:study_quants_flash}
\end{figure*}

\begin{figure*}[tbp]
  \centering
  \begin{subfigure}[b]{\resultswidth}
    \includegraphics[width=\textwidth]{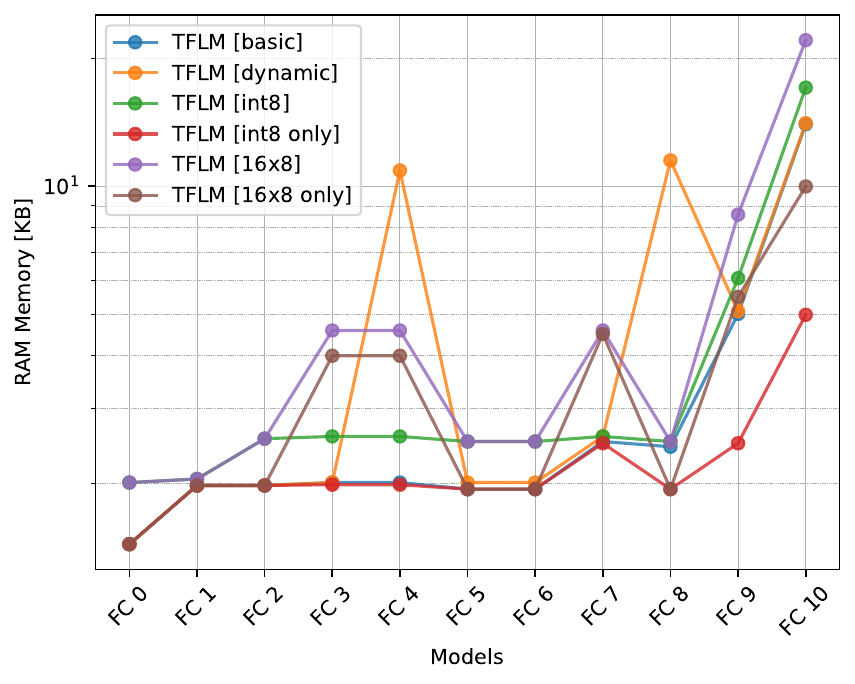}
    \label{fig:study_quants_fc_ram}
  \end{subfigure}
  \begin{subfigure}[b]{\resultswidth}
    \vskip 5mm  
    \includegraphics[width=\textwidth]{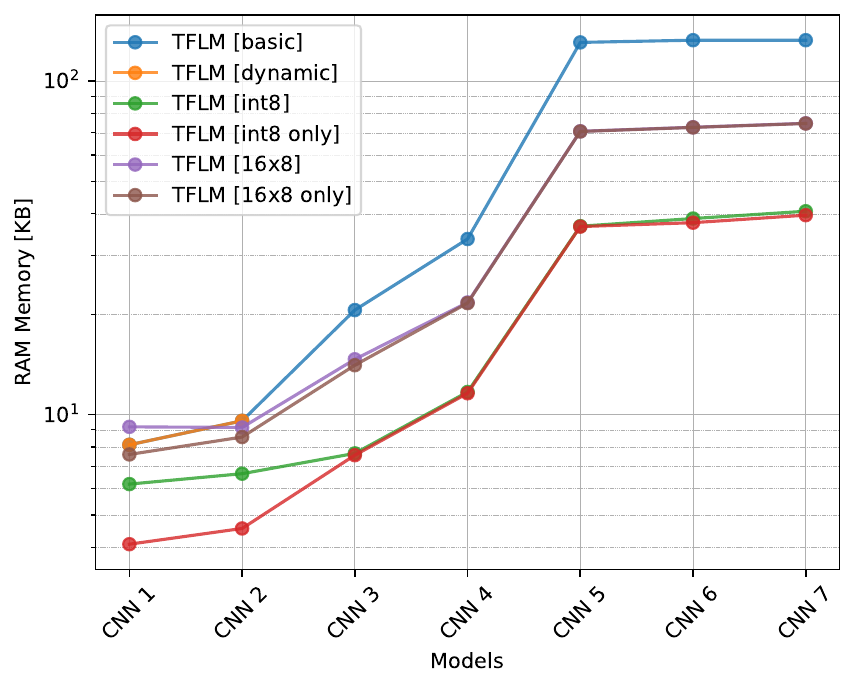}
    \label{fig:study_quants_cnn_ram}
  \end{subfigure}
  \caption{RAM usage across various quantization schemes. The models were tested on NUCLEO-L4R5ZI.}
  \label{fig:study_quants_ram}
\end{figure*}

To perform the evaluation, we tested the \gls{fc} and \gls{cnn} models on the two available boards. Figures~\ref{fig:study_quants_error}--\ref{fig:study_quants_ram} compare these setups in terms of deployment error, execution time, flash size, and RAM usage on the STM board.
\begin{itemize}
  \item \textbf{Model correctness:}
  \begin{itemize}
    \item We recommend avoiding using \textit{dynamic} quantization since it lacks proper support (evidenced by missing data points and high error rates in Fig.~\ref{fig:study_quants_error}). Models using this scheme often fail to run or exhibit unacceptably high errors. Even when deployment succeeds, it does not outperform other quantization schemes in any metric.
    \item During this and some other experiments, some models could not be executed successfully on the Renesas board, with the program unexpectedly stopping during execution. The root cause of this behavior remains unclear.
    \item Fig.~\ref{fig:study_quants_error} shows that all other quantization schemes reach an acceptable error. The \textit{basic} conversion perfectly produces the expected outputs, while other types of quantization introduce a bit of error rooted in the nature of quantization. Between \textit{int8} and \textit{16x8} quantization schemes, as expected, \textit{16x8} shows smaller error rates due to its higher precision representation of activations.
  \end{itemize}
  \item \textbf{Execution time:} Although, based on the results illustrated in Fig.~\ref{fig:study_quants_exe}, determining the fastest quantization scheme may not be clear, the following statements appear to summarize the key points. For \gls{fc} models, the \textit{int8 only} and \textit{16x8 int only} quantizations outperform their non-only counterparts, with \textit{16x8 int only} performing best for smaller models and \textit{int8 only} excelling for larger models. For \gls{cnn} models, \textit{basic} and \textit{dynamic} quantizations are significantly slower due to their lack of CMSIS-NN library optimizations. The \textit{int8} and \textit{16x8} variants have similar performance, with \textit{16x8 int only} slightly outperforming on smaller networks and \textit{int8 only} proving better for larger ones.
  \item \textbf{Flash size:} As shown in Fig.~\ref{fig:study_quants_flash}, all quantization schemes start with relatively similar flash sizes, but as the model scales, the \textit{basic} version's flash size increases fourfold (four bytes per parameter), whereas the other variants increase by only one byte per parameter. Therefore, the \textit{basic} model gets worse as model size increases.
  \item \textbf{RAM usage:} Based on Fig.~\ref{fig:study_quants_ram}, for \gls{cnn} models, as expected, the \textit{int8}-based variants require less RAM than other types of the same model, since they store activations in 8-bit representation. These are followed by \textit{16x8}-based models, and finally, \textit{basic} and \textit{dynamic} schemes requiring the most RAM. For \gls{fc} models, RAM consumption patterns are less predictable, though the \textit{int8 only} scheme consistently proves efficient, often matching or outperforming other methods.
  \item \textbf{Conclusion:} The choice of quantization scheme depends on many factors and can lead to execution that is a couple of times faster and requires less memory. However, \textit{int8 only} quantization is a good choice for most cases. Together with the \textit{basic} (non-quantized) variant, these two schemes will serve as the default choices for the remainder of this study.
\end{itemize}

\subsubsection{Pruning and Clustering}
This study examines the effects of pruning and clustering on model performance. After quantization, pruning is the most popular technique for reducing the size of a model. However, only a few sources make it clear that unstructured pruning will not bring any benefit to general-purpose processors. To leverage this technique, either specialized hardware is required, or pruning must be structured--eliminating entire units of the model, such as neurons or channels, thereby creating a model with a modified architecture.

The counterintuitive nature of this limitation lies in the fact that, in unstructured pruning, the zero values for the pruned weights still occupy memory and consume the same storage space as non-pruned weights. It is worth noting that pruned models can be compressed more effectively due to the repetitive zero values in their weights. However, such compression requires decompression before execution, which negates most practical gains. Moreover, in terms of runtime performance, multiplying a pruned weight (zero) with another value takes the same clock cycles as a multiplication involving non-zero weights. Consequently, unstructured pruning yields no improvement in execution time. Similarly, clustering, which involves grouping weights into discrete clusters, offers little to no advantage in improving the model's performance on general-purpose hardware.

\begin{figure*}[tbp]
  \centering
  \begin{subfigure}[b]{\resultswidth}
    \includegraphics[width=\textwidth]{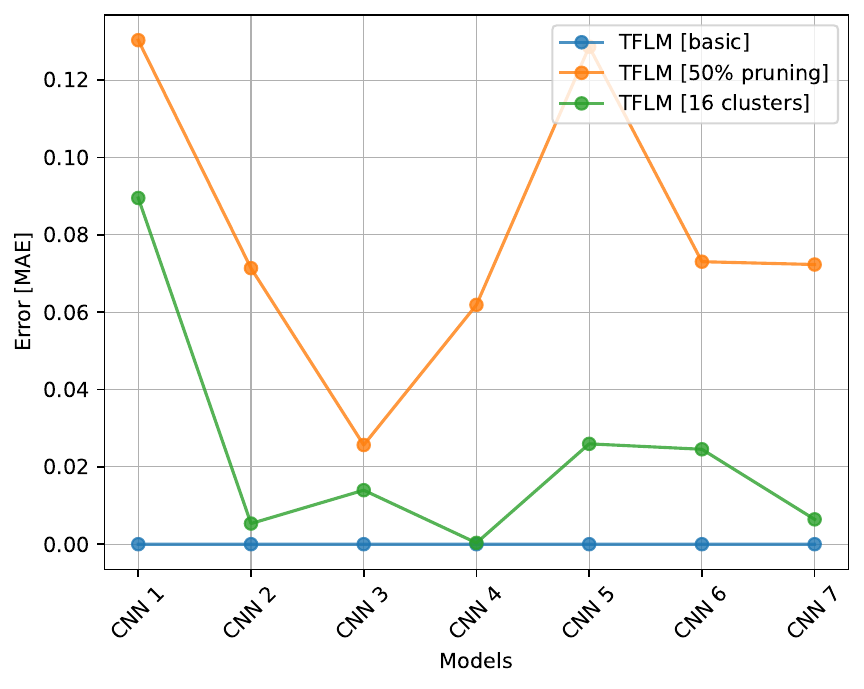}
    \caption{Deployment error.}
    \label{fig:study_pruning_clustering_cnn_error}
  \end{subfigure}
  \begin{subfigure}[b]{\resultswidth}
    \vskip 3mm  
    \includegraphics[width=\textwidth]{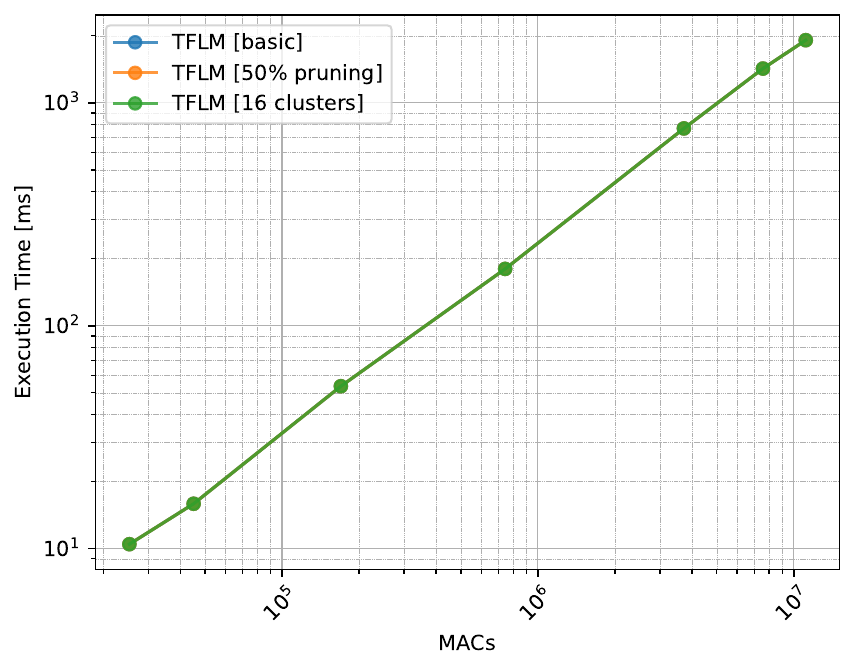}
    \caption{Execution time. The three plots are drawn on top of each other, appearing as one.}
    \label{fig:study_pruning_clustering_cnn_exe}
  \end{subfigure}
  \caption{Performance evaluation of basic, pruned, and clustered \gls{cnn} models on the NUCLEO-L4R5ZI board.}
  \label{fig:study_pruning_clustering}
\end{figure*}

To elucidate these findings, we evaluated \gls{fc} and \gls{cnn} models on the NUCLEO-L4R5ZI board. In our experiments, we applied 50\% sparsity for pruning and clustering with 16 centroids. These techniques increase the error of the models (see Fig.~\ref{fig:study_pruning_clustering_cnn_error}), which can be mitigated to some extent through fine-tuning. However, as illustrated in Fig.\ref{fig:study_pruning_clustering_cnn_exe}, the execution times for the basic, pruned, and clustered models are nearly identical, with less than 0.01\% difference. This negligible variation means the plots are essentially overlaid. The same holds true for the flash memory usage and RAM requirements of the models.

\subsubsection{TFLM vs. Edge Impulse}
\Gls{tflm} and Edge Impulse are two prominent tools in the \gls{eai} domain. While Edge Impulse originally relied on \gls{tflm} and still provides the option to deploy models using it, it has significantly evolved over the years. With features like its EON compiler, Edge Impulse introduces a unique approach to model execution. However, it is worth noting that Edge Impulse still utilizes \gls{tflm}'s core kernels in most cases.

\begin{figure*}[tbp]
  \centering
  \begin{subfigure}[b]{\resultswidth}
    \includegraphics[width=\textwidth]{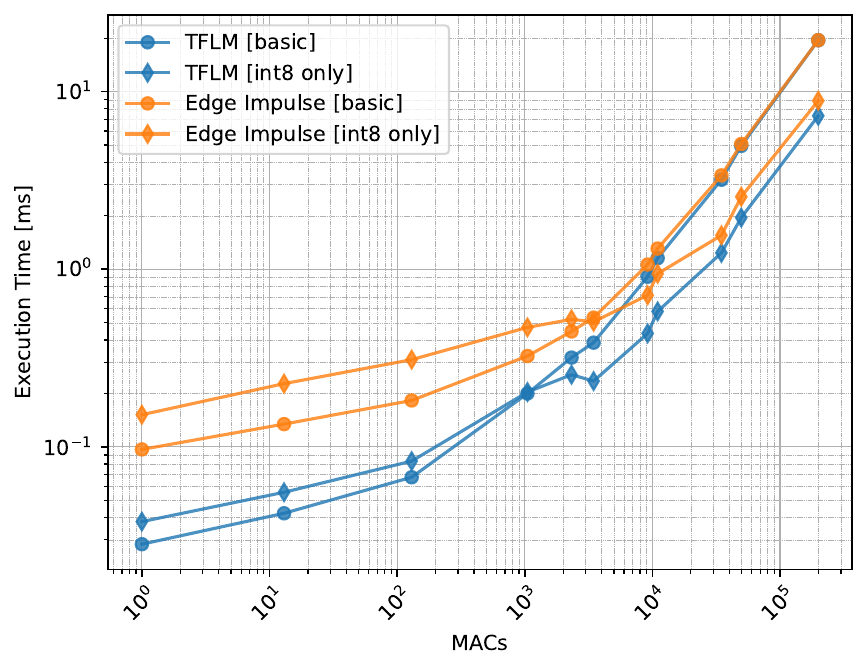}
    \caption{Execution time.}
    \label{fig:study_ei_fc_exe}
  \end{subfigure}
  \begin{subfigure}[b]{\resultswidth}
    \vskip 5mm  
    \includegraphics[width=\textwidth]{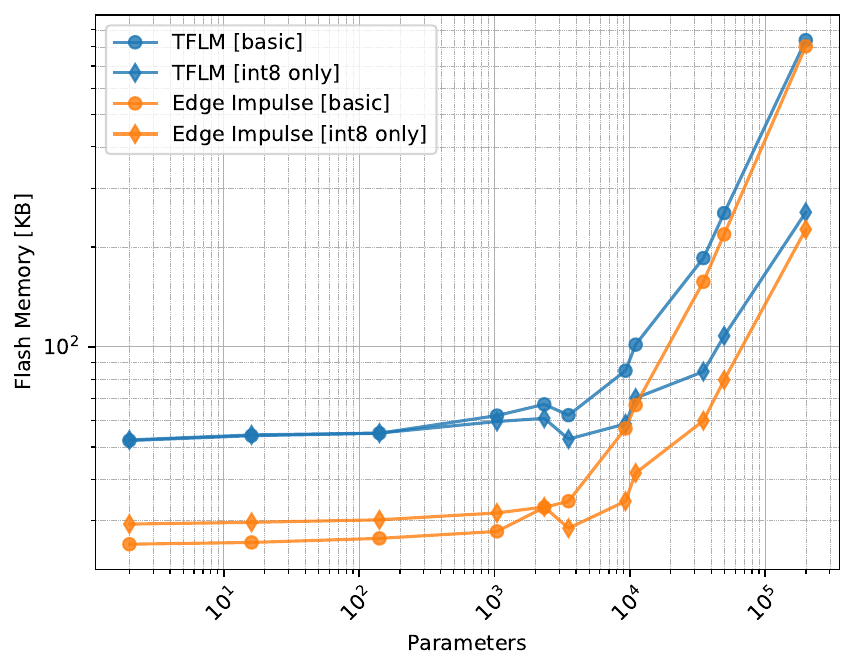}
    \caption{Flash size.}
    \label{fig:study_ei_fc_flash}
  \end{subfigure}
  \caption{Performance evaluation of \gls{fc} models deployed using Edge Impulse and \gls{tflm} on the STM board.}
  \label{fig:study_ei}
\end{figure*}

As Edge Impulse does not natively support \glspl{rnn}, this study focuses on testing \gls{fc}, \gls{cnn}, and MLPerf Tiny models deployed on the STM and Renesas boards. The results show that on the Renesas board, \gls{tflm} outperforms Edge Impulse in terms of execution time, flash size, and RAM usage. On the STM board, the key findings are as follows:
\begin{itemize}
  \item \textbf{Model correctness:} Since the MBNet model from the MLPerf Tiny test suite was too large in most test cases, it has been excluded from the comparison. Quantized models introduced a minor degree of error in the outputs, though this error was negligible and within acceptable margins. Notably, Edge Impulse's models were more accurate in producing the expected outputs compared to \gls{tflm}.
  \item \textbf{Execution time:} \Gls{tflm} demonstrated faster execution times for smaller \gls{fc} models. For other models, the two were in par, with a slight edge for \gls{tflm}.
  \item \textbf{Flash size:} Edge Impulse consumed less flash memory compared to \gls{tflm}. The extent of this difference varied based on the model's type and size, ranging from negligible to over 100\% improvement. The differentiating factor is the library's base size, and the amount of improvement diminishes as the model grows larger.
  \item \textbf{RAM usage:} For small to medium-sized \gls{fc} models, Edge Impulse showed better RAM efficiency. For other model types, the memory usage of both libraries was nearly equivalent.
  \item \textbf{Conclusion:} On the Renesas RX65N board, \gls{tflm} is the preferred tool for deployment. On the NUCLEO-L4R5ZI board, \gls{tflm} is recommended when execution speed is a priority, while Edge Impulse may be more suitable in applications where flash size or RAM usage is more critical.
\end{itemize}

\subsubsection{TFLM vs. Ekkono}
Ekkono is a lightweight edge AI library designed to support incremental learning directly on embedded devices. Since Ekkono is not able to do classification, we have slightly changed some models to make them suitable for regression. The changes are minimal and should not have a noticeable impact on the results.

\begin{figure*}[tbp]
  \centering
  \begin{subfigure}[b]{\resultswidth}
    \includegraphics[width=\textwidth]{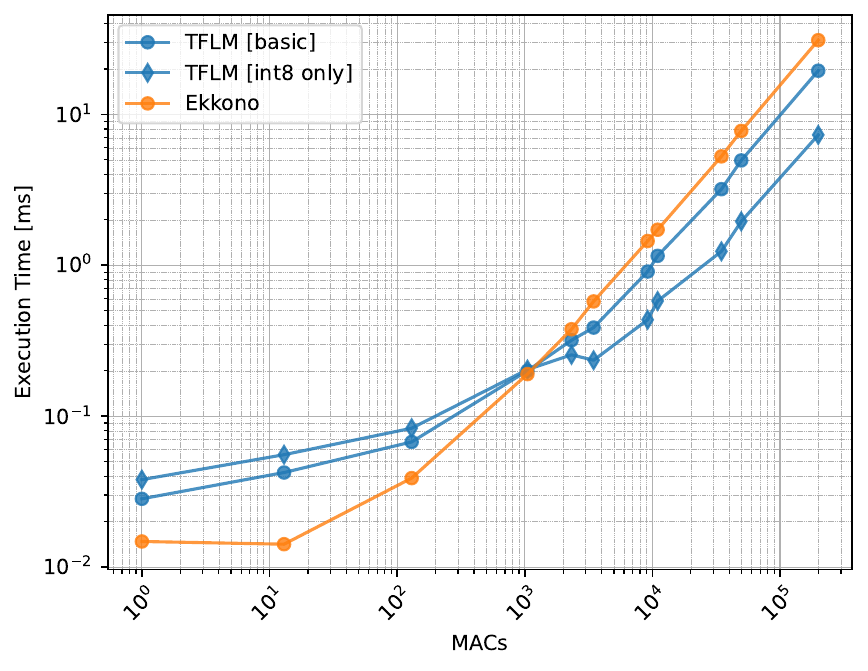}
    \caption{Execution time.}
    \label{fig:study_ekkono_exe}
  \end{subfigure}
  \begin{subfigure}[b]{\resultswidth}
    \vskip 5mm  
    \includegraphics[width=\textwidth]{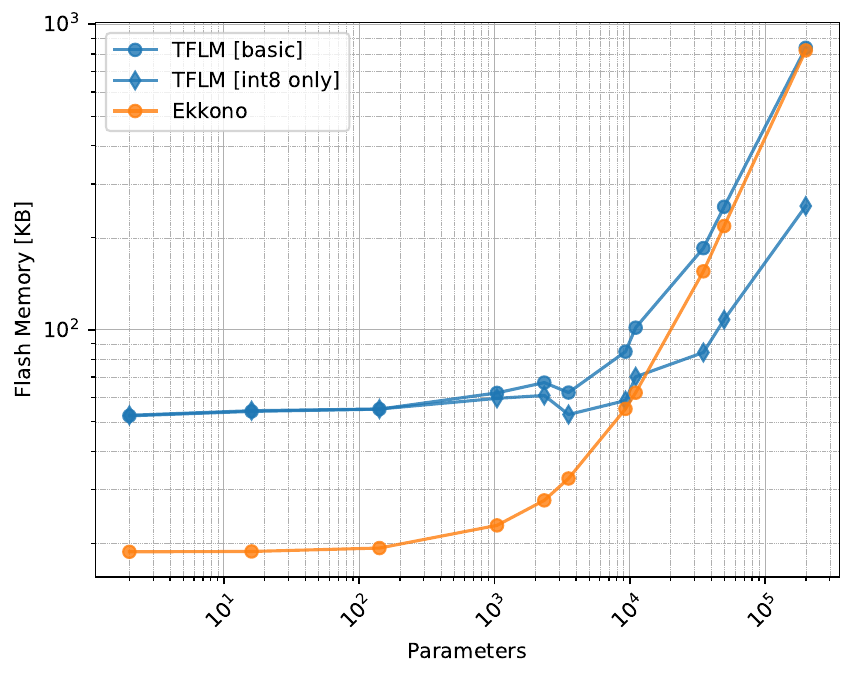}
    \caption{Flash size.}
    \label{fig:study_ekkono_flash}
  \end{subfigure}
  \caption{Performance evaluation of Ekkono and \gls{tflm} models on the NUCLEO-L4R5ZI board.}
  \label{fig:study_ekkono}
\end{figure*}

\begin{itemize}
  \item \textbf{Model correctness:} Both Ekkono and \gls{tflm}'s \textit{basic} version can perfectly reproduce the expected values. \gls{tflm}'s \textit{int8 only} version has a bit of error which remains in the acceptable range.
  \item \textbf{Execution time:} As shown in Fig.~\ref{fig:study_ekkono_exe}, Ekkono delivered faster execution times for smaller models. In contrast, \gls{tflm} performed better with larger models, particularly when using the \textit{int8 only} version.
  \item \textbf{Flash size:} As illustrated in Fig.~\ref{fig:study_ekkono_flash}, Ekkono, being a simpler and more lightweight library, offers a smaller baseline flash memory footprint than \gls{tflm} (18.8 kB vs. 52.2 kB). However, as model size increases, the performance gap narrows, with the \textit{basic} version of \gls{tflm} catching up. This is while the \textit{int8 only} version of \gls{tflm} surpases their performance and becomes the best choice for large models.
  \item \textbf{RAM usage:} Similar to flash size trends, Ekkono is more efficient with smaller models, while \gls{tflm}'s \textit{int8 only} variant performs better as models grow in size. Nevertheless, the memory requirements for all tested \gls{fc} models were relatively modest (maximum 14 kB) and are unlikely to constrain most applications.
  \item \textbf{Conclusion:} For smaller models, Ekkono provides a more efficient solution. However, as model size increases, \gls{tflm}'s \textit{int8 only} variant emerges as the more effective choice.
\end{itemize}

\subsubsection{TFLM vs. eAI Translator}
Renesas eAI Translator is a vendor-specific tool designed to enable the execution of neural networks on Renesas microcontrollers. Our testing hardware, the Renesas RX65N, incorporates a proprietary RXv2 CPU architecture. Since \gls{tflm} lacks optimized kernel implementations for RXv2, it handles this architecture as a general-purpose CPU. In contrast, Renesas eAI Translator provides the capability to fine-tune implementations specifically for the target hardware, potentially achieving better performance. In this study, we assess whether such hardware-specific tuning allows eAI Translator to outperform the more generic \gls{tflm}.

\begin{figure*}[tbp]
  \centering
  \begin{subfigure}[b]{\resultswidth}
    \includegraphics[width=\textwidth]{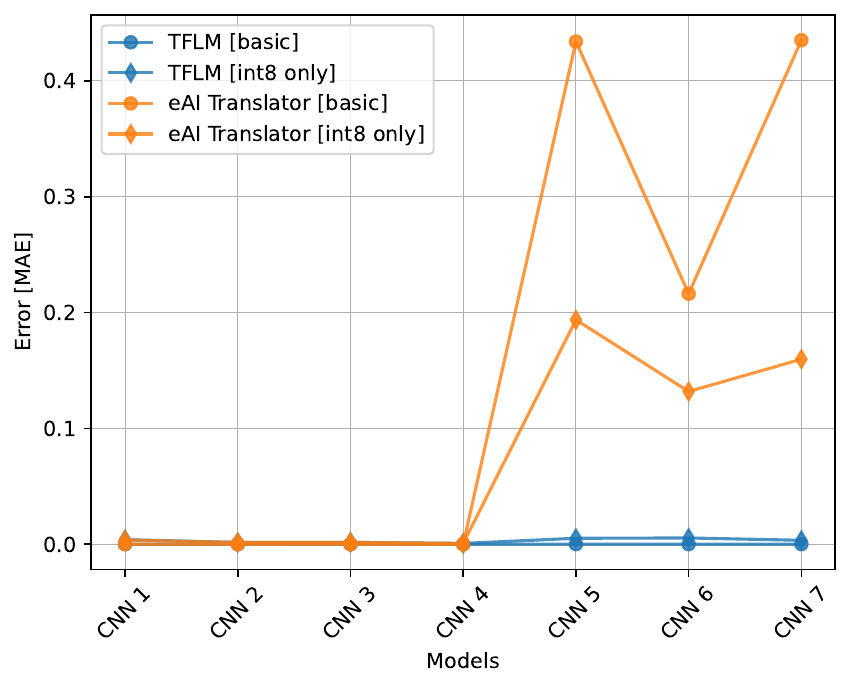}
    \caption{Deployment error.}
    \label{fig:study_translator_cnn_error}
  \end{subfigure}
  \begin{subfigure}[b]{\resultswidth}
    \vskip 5mm  
    \includegraphics[width=\textwidth]{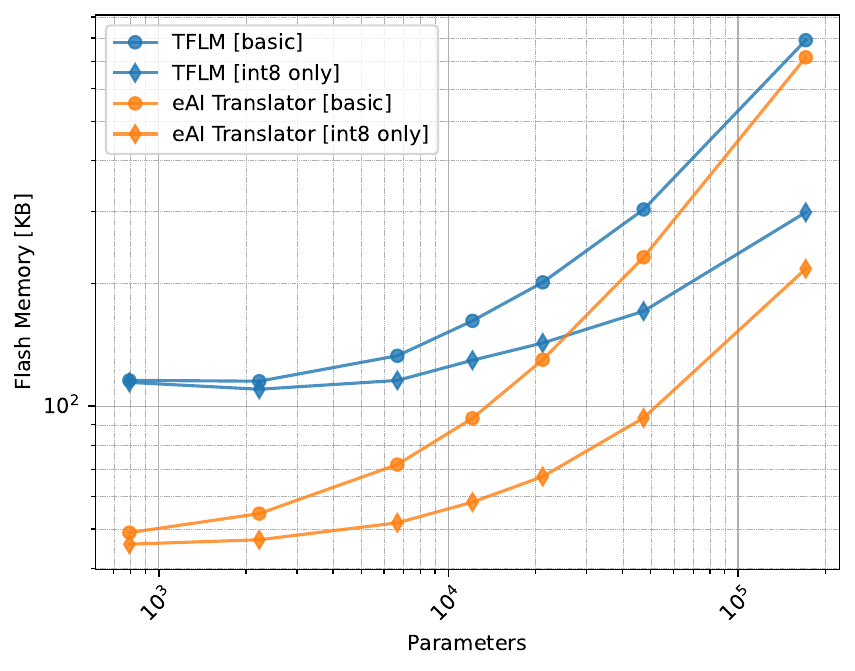}
    \caption{Flash size.}
    \label{fig:study_translator_cnn_flash}
  \end{subfigure}
  \caption{Performance evaluation of \gls{cnn} models deployed using eAI Translator and \gls{tflm} on the Renesas RX65N board.}
  \label{fig:study_translator}
\end{figure*}

\begin{itemize}
  \item \textbf{Model correctness:} The error rate of \gls{tflm} and eAI Translator are normally the same, but for some large models, \gls{tflm} is more suitable due to lower error rates.
  \item \textbf{Execution time:} Renesas has optimized the CMSIS-NN library to align with the hardware specifications of RX microcontrollers. We utilized this optimized library with eAI Translator but not with \gls{tflm}. As a result, we exclude the \textit{int8 only} variants of both platforms from the execution time comparison. Focusing on their \textit{basic} variants, our results show that eAI Translator generally outperforms \gls{tflm}, particularly on smaller models.
  \item \textbf{Flash size:} The eAI Translator library requires significantly less flash memory compared to \gls{tflm}. This makes eAI Translator a stronger candidate for use in memory-constrained environments, especially when dealing with smaller models.
  \item \textbf{RAM usage:} In terms of RAM consumption, eAI Translator demonstrates slightly better efficiency than \gls{tflm}.
  \item \textbf{Conclusion:} Overall, except in a few cases highlighted above, eAI Translator is often the better choice across all evaluated metrics, including execution time, memory usage, and flash size.
\end{itemize}

\section{Conclusion}
\label{sec:conclusion}

This paper explores the landscape of \gls{eai} tools, introduces an automation system to streamline their usage, and benchmarks their performance across various metrics. The review identifies key features of these tools, revealing that while they generally support techniques like quantization, many advanced methods in the field remain absent.

To address the challenges in evaluating \gls{eai} tools, we developed EdgeMark, an automation system designed to simplify model deployment workflows. EdgeMark provides a unified interface for deploying models on target devices, automating the entire process of model generation, optimization, conversion, and deployment. The system is user-friendly, modular, and easily extensible, enabling researchers and developers to test their ideas with minimal effort.

The benchmarking results showed that the performance of \gls{eai} tools varies across different metrics. The choice of quantization scheme, for example, can significantly impact execution time, flash size, and RAM usage. The study also compared the performance of \gls{tflm} with Edge Impulse, Ekkono, and Renesas eAI Translator. The results showed that the choice of tool depends on the application requirements, with each tool having its own strengths and weaknesses.

The insights from this work can guide developers in selecting the most suitable \gls{eai} tools based on their specific application requirements, while the automation system lowers the adoption barrier for such technologies. Future work will focus on expanding the benchmark to include more tools and platforms, as well as incorporating energy measurements to provide a more comprehensive evaluation of \gls{eai} tools. An interesting area for further exploration is evaluating these tools on hardware accelerators specifically designed for neural network processing. Further, the automation system can be used in \gls{nas} research to provide real measurements of the performance of generated models.

As embedded systems increasingly rely on \gls{ai}-driven capabilities, this study provides a foundational step toward creating streamlined workflows and taking informed decisions in selecting and deploying \gls{eai} tools, ultimately advancing the integration of \gls{ai} into resource-constrained devices.

\section*{Acknowledgements}
This work was supported by the Innovation Fund Denmark for the Project Digital Research Centre Denmark (DIREC) under Grant 9142-00001B.

\appendix
\section{Additional Experiments}
\label{sec:appendix}

\subsection{RNNs}

\begin{figure*}[tbp]
  \centering
  \begin{subfigure}[b]{\resultswidth}
    \includegraphics[width=\textwidth]{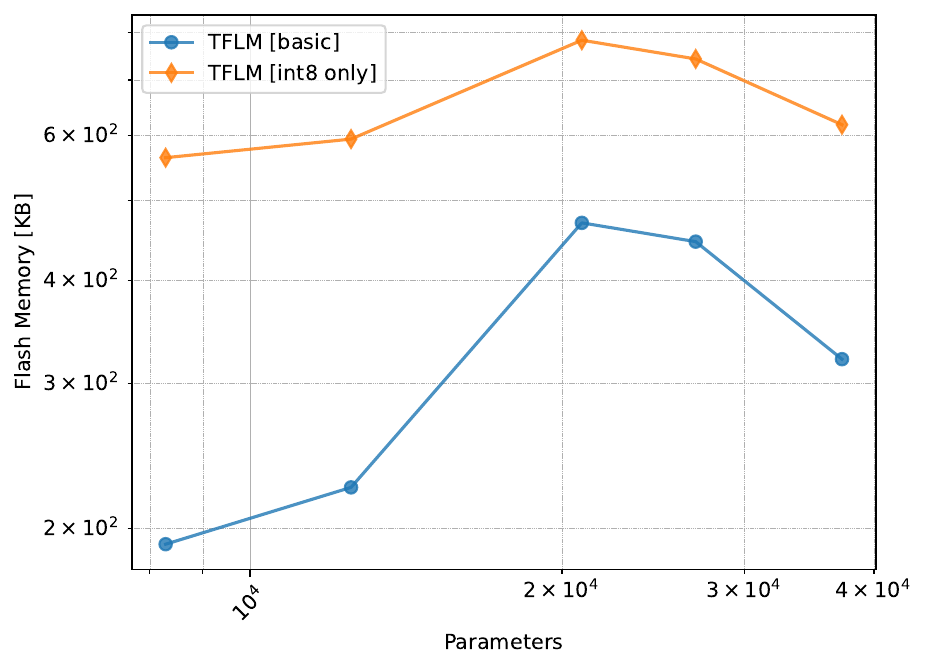}
    \caption{Flash size.}
    \label{fig:study_rnn_flash}
  \end{subfigure}
  \begin{subfigure}[b]{\resultswidth}
    \vskip 5mm  
    \includegraphics[width=\textwidth]{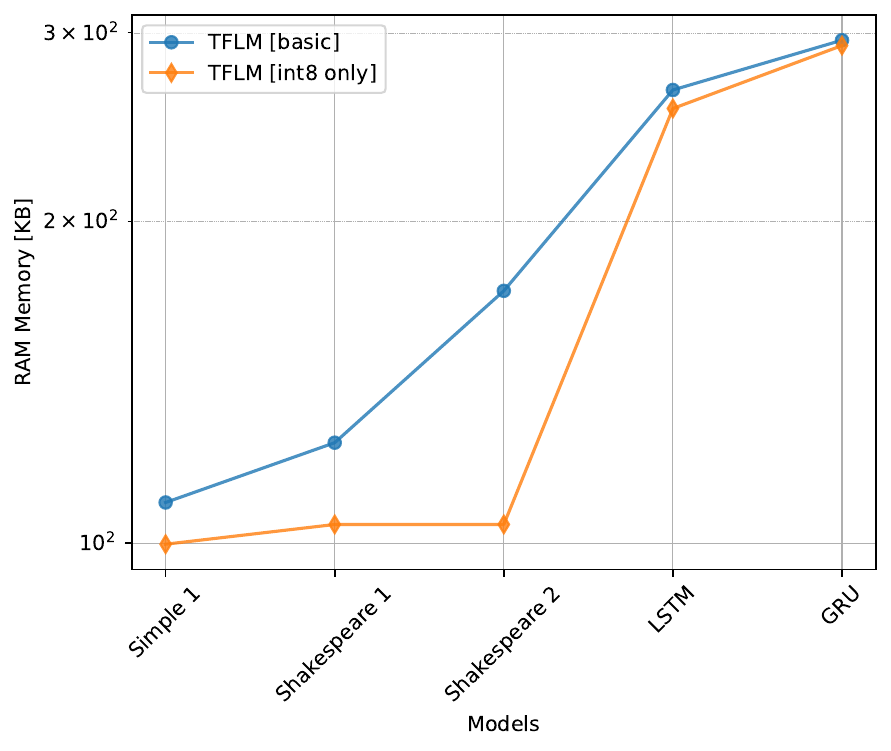}
    \caption{RAM usage.}
    \label{fig:study_rnn_ram}
  \end{subfigure}
  \caption{Performance evaluation of RNN models on the NUCLEO-L4R5ZI board.}
  \label{fig:study_rnn}
\end{figure*}

\Gls{rnn} models are widely used in applications that require processing sequential data. However, deploying these resource-intensive models on embedded devices requires careful consideration. In this study, we evaluated the performance of SimpleRNN, LSTM, and GRU models using \gls{tflm} on the NUCLEO-L4R5ZI board. The models are introduced in Section \ref{sec:models}, and the results can be found in Table~\ref{tab:RNN_results}. To provide a clear comparison, memory usage is visualized in Fig.\ref{fig:study_rnn}, from which \textit{Simple 0} is excluded due to its negligible resource requirements compared to other models. Similarly, \textit{Simple 2} is excluded because its \textit{basic} version failed to execute on the board.

\begin{table*}[htbp]
  \caption{Performance Evaluation of \gls{rnn} Models}
  \begin{center}
    \begin{tabular}{|l|l|l|l|l|l|}
      \hline
      \textbf{Model} & \textbf{Variant} & \textbf{Error} & \textbf{Exe {\scriptsize (ms)}} & \textbf{Flash {\scriptsize (kB)}} & \textbf{RAM {\scriptsize (kB)}} \\
      \hline
      \multirow{2}{*}{Simple 0} & basic & 0 & 0.11 & 110.4 & 8.1 \\ \cline{2-6}
      & int8 only & 0.006 & 0.14 & 111.6 & 7.1 \\
      \hline
      \multirow{2}{*}{Simple 1} & basic & 0 & 107.2 & 218.7 & 112.6 \\ \cline{2-6}
      & int8 only & 0.004 & 292.9 & 590.8 & 103.3 \\
      \hline
      \multirow{2}{*}{Simple 2} & basic & - & - & - & - \\ \cline{2-6}
      & int8 only & 0.004 & 292.9 & 615.2 & 106.4 \\
      \hline
      \multirow{2}{*}{Shakespeare 1} & basic & 0 & 141.2 & 251.6 & 127.6 \\ \cline{2-6}
      & int8 only & 0.021 & 168.4 & 620.8 & 107.6 \\
      \hline
      \multirow{2}{*}{Shakespeare 2} & basic & 0 & 377.2 & 348.2 & 175.6 \\ \cline{2-6}
      & int8 only & 0.022 & 323.5 & 645.2 & 107.6 \\
      \hline
      \multirow{2}{*}{LSTM} & basic & 0 & 362.2 & 473.0 & 268.7 \\ \cline{2-6}
      & int8 only & 0.014 & 565.3 & 769.8 & 258.4 \\
      \hline
      \multirow{2}{*}{GRU} & basic & 0 & 276.7 & 497.0 & 298.7 \\ \cline{2-6}
      & int8 only & 0.044 & 374.4 & 809.8 & 295.4 \\
      \hline
    \end{tabular}
    \label{tab:RNN_results}
  \end{center}
\end{table*}

\begin{itemize}
  \item \textbf{Model correctness:} With the exception of the \textit{basic} version of \textit{Simple 2} which runs out of memory, all models achieved acceptable error rates. The most concerning case is \textit{GRU}, which may require more attention.
  \item \textbf{Execution time:} Surprisingly, the \textit{basic} variants of the models show faster execution times than their \textit{int8 only} counterparts, with the exception of \textit{Shakespeare 2}.
  \item \textbf{Flash size:} Against our expectations, Fig.~\ref{fig:study_rnn_flash} reveals that the \textit{int8 only} variants require significantly more flash memory compared to the \textit{basic} versions.
  \item \textbf{RAM usage:} As shown in Fig.~\ref{fig:study_rnn_ram}, the \textit{int8 only} variants use less RAM than \textit{basic} models. However, we expected them to have the advantage with a bigger margin.
  \item \textbf{Conclusion:} This study had many unexpected results. Notably, the \textit{int8 only} variants of the models are not as efficient as expected. The \textit{basic} versions are faster and consume less flash memory, while they require slightly more RAM.
\end{itemize}

It is believed that LSTM is more powerful than GRU, while GRU is often favored for its computational and parameter efficiency. Consistent with this, our study observes that similar LSTMs have more parameters and a higher number of \glspl{mac} compared to GRUs. However, it is interesting that using TFLM, GRUs require more flash memory and RAM than LSTMs. Despite this, GRUs still manage to execute faster than LSTMs.

\subsection{GCC Optimization Levels}

\begin{figure*}[tbp]
  \centering
  \begin{subfigure}[b]{\resultswidth}
    \includegraphics[width=\textwidth]{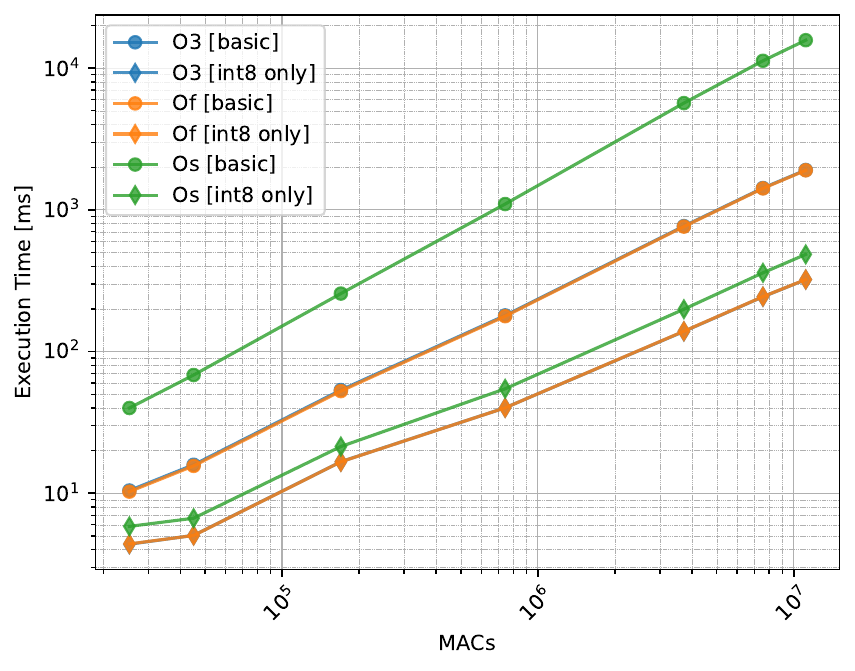}
    \caption{Execution time. \textit{O3} and \textit{Of} are almost overlapped.}
    \label{fig:study_opt_cnn_exe}
  \end{subfigure}
  \begin{subfigure}[b]{\resultswidth}
    \vskip 5mm  
    \includegraphics[width=\textwidth]{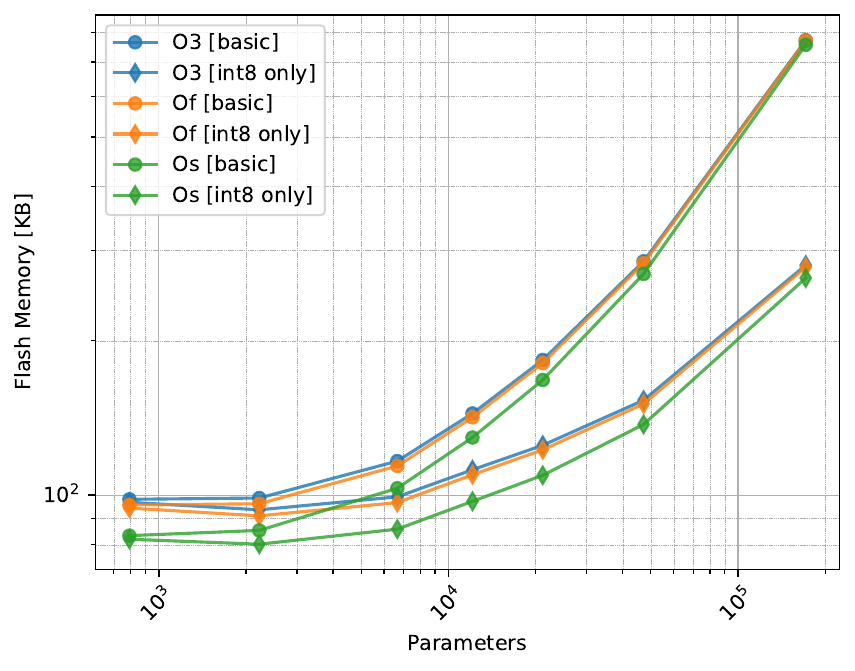}
    \caption{Flash size. \textit{O3} and \textit{Of} are almost overlapped.}
    \label{fig:study_opt_cnn_flash}
  \end{subfigure}
  \caption{Performance evaluation of \gls{cnn} models with different GCC optimization levels on the NUCLEO-L4R5ZI board.}
  \label{fig:study_opt}
\end{figure*}

The \gls{gcc} is a widely used compiler for C and C++ programming languages. \Gls{gcc} provides several optimization levels, ranging from \textit{O0} (no optimization) to \textit{O3} (maximum optimization). In this study, we evaluated the impact of \textit{Os} (optimize for size), \textit{Of} (optimize for speed), and \textit{O3} on the performance of \gls{fc} and \gls{cnn} models deployed on the NUCLEO-L4R5ZI board using \gls{tflm}.

\begin{itemize}
  \item \textbf{Model correctness:} All three optimization levels produce the exact same outputs, confirming that the correctness of the models remains unaffected by these optimizations.
  \item \textbf{Execution time:} As shown in Fig.~\ref{fig:study_opt_cnn_exe}, the \textit{O3} and \textit{Of} optimization levels perform almost identically and better than \textit{Os}.
  \item \textbf{Flash size:} The \textit{O3} and \textit{Of} optimizations resulted in similar flash memory usage, while \textit{Os} produced a slightly smaller flash size.
  \item \textbf{RAM usage:} No notable differences in RAM usage were observed across the three optimization levels.
  \item \textbf{Conclusion:} Both \textit{O3} and \textit{Of} provide comparable performance and outperform \textit{Os} in terms of execution time. However, if reducing flash size is a priority, \textit{Os} offers a slight advantage.
\end{itemize}

\subsection{Importance of FPU}

The \gls{fpu} is a specialized coprocessor designed to handle floating-point arithmetic operations efficiently. To check the speedup provided by the \gls{fpu}, we evaluated the performance of \gls{fc} and \gls{cnn} models on the NUCLEO-L4R5ZI board with and without the \gls{fpu} enabled.

\begin{figure*}[tbp]
  \centering
  \begin{subfigure}[b]{\resultswidth}
    \includegraphics[width=\textwidth]{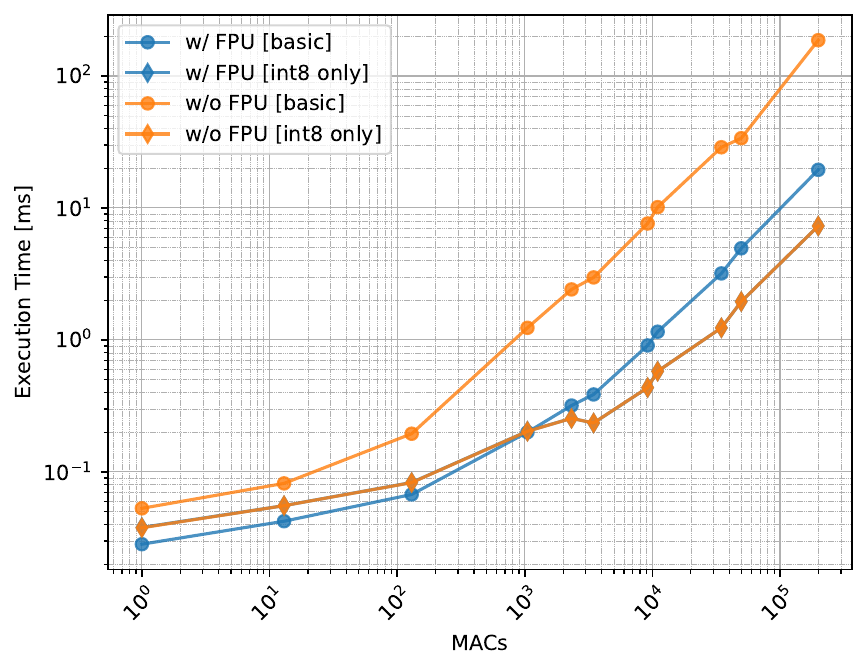}
    \caption{\Gls{fc} models. \textit{int8 only} variants are overlapped.}
    \label{fig:study_fpu_fc_exe}
  \end{subfigure}
  \begin{subfigure}[b]{\resultswidth}
    \vskip 5mm  
    \includegraphics[width=\textwidth]{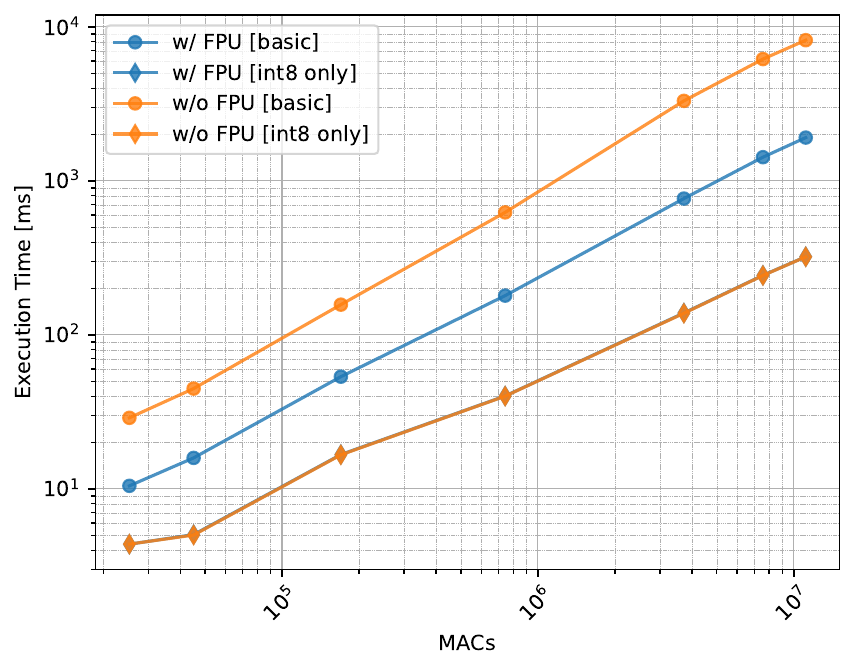}
    \caption{\Gls{cnn} models. \textit{int8 only} variants are overlapped.}
    \label{fig:study_fpu_cnn_exe}
  \end{subfigure}
  \caption{Execution time of models with and without the \gls{fpu} on the NUCLEO-L4R5ZI board.}
  \label{fig:study_fpu}
\end{figure*}

Our results show that \gls{fpu} does not have a noticable impact on any of the evaluated metrics other than execution time. As illustrated in Fig.~\ref{fig:study_fpu}, the \gls{fpu} provides a significant speedup (85\% to 850\%) for the \textit{basic} variants of both \gls{fc} and \gls{cnn} models. Still, if using \textit{int8 only}, utilizing the CMSIS-NN library can be even more beneficial and there is no need for the \gls{fpu}.

\subsection{STM vs. Renesas}

STMicroelectronics and Renesas are two major semiconductor companies that provide microcontrollers for embedded systems. STMicroelectronics is more popular between researchers and hobbyists, while Renesas is more focused on industrial and automotive applications.

\begin{figure*}[tbp]
  \centering
  \begin{subfigure}[b]{\resultswidth}
    \includegraphics[width=\textwidth]{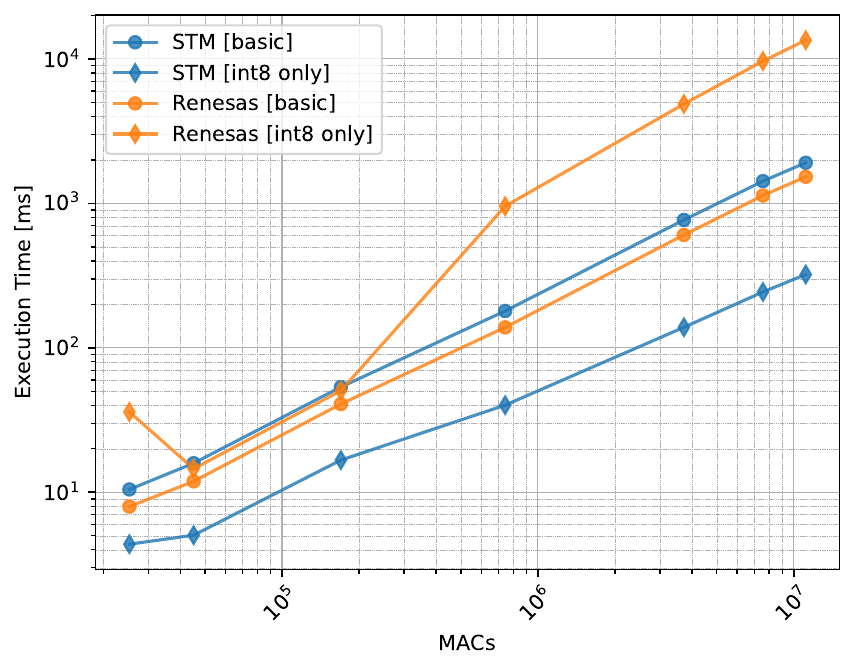}
    \caption{Execution time. The \textit{int8 only} version of the Renesas board does not utilize CMSIS-NN and is not comparable to its STM counterpart.}
    \label{fig:study_stm_renesas_cnn_exe}
  \end{subfigure}
  \begin{subfigure}[b]{\resultswidth}
    \vskip 5mm  
    \includegraphics[width=\textwidth]{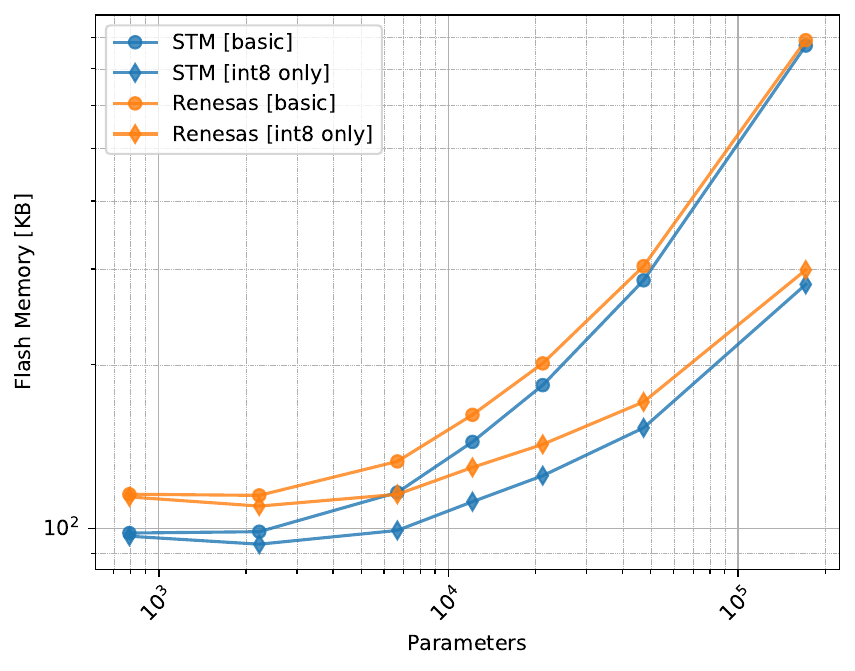}
    \caption{Flash size.}
    \label{fig:study_stm_renesas_cnn_flash}
  \end{subfigure}
  \caption{Performance evaluation of \gls{cnn} models on the STM and Renesas boards.}
  \label{fig:study_stm_renesas}
\end{figure*}

The two selected boards from these companies, STM NUCLEO-L4R5ZI and Renesas RX65N, have similar specifications, including identical flash size, RAM, and clock speed, along with many other shared features. However, the RX65N board has a proprietary RXv2 CPU architecture, while the NUCLEO-L4R5ZI board uses an ARM Cortex-M4 core. This distinction makes the comparison between these two boards particularly intriguing.

\Gls{fc}, \gls{cnn}, \gls{rnn}, and MLPerf Tiny models were evaluated on both boards using \gls{tflm}. It is worth noting that these boards have various configurable settings, which may favor one platform over the other in specific scenarios. Our study primarily relied on default settings, however it should not be considered as a comprehensive comparison between the two boards.

\begin{itemize}
  \item \textbf{Model correctness:} The Renesas board fails to run a few of the models. Aside from this, the two boards provide similar results.
  \item \textbf{Execution time:} When evaluating the \textit{basic} versions of models and excluding \textit{int8-only} variants (since CMSIS-NN optimizations are used only on the STM board), the Renesas board performs slightly better in terms of execution time. This is illustrated in Fig.~\ref{fig:study_stm_renesas_cnn_exe}.
  \item \textbf{Flash size:} As shown in Fig.~\ref{fig:study_stm_renesas_cnn_flash}, the STM board requires less flash memory compared to the Renesas board. The difference is more noticeable for smaller models. This is mainly due to the difference between the size of the two libraries, which becomes less significant as the model size increases.
  \item \textbf{RAM usage:} The RAM usage of the models is almost identical across the two boards.
  \item \textbf{Conclusion:} The two boards seem to have a relatively similar performance. The Renesas board might be slightly faster, but the STM board has a bit smaller flash size.
\end{itemize}

\subsection{GCC vs CCRX}

\begin{figure*}[tbp]
  \centering
  \begin{subfigure}[b]{\resultswidth}
    \includegraphics[width=\textwidth]{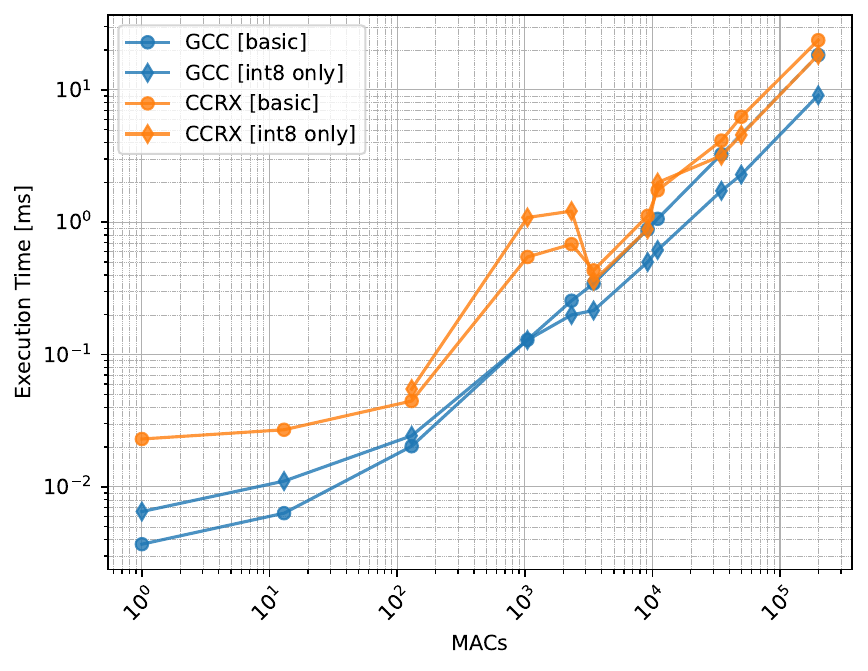}
    \caption{\Gls{fc} models.}
    \label{fig:study_ccrx_fc_exe}
  \end{subfigure}
  \begin{subfigure}[b]{\resultswidth}
    \vskip 5mm  
    \includegraphics[width=\textwidth]{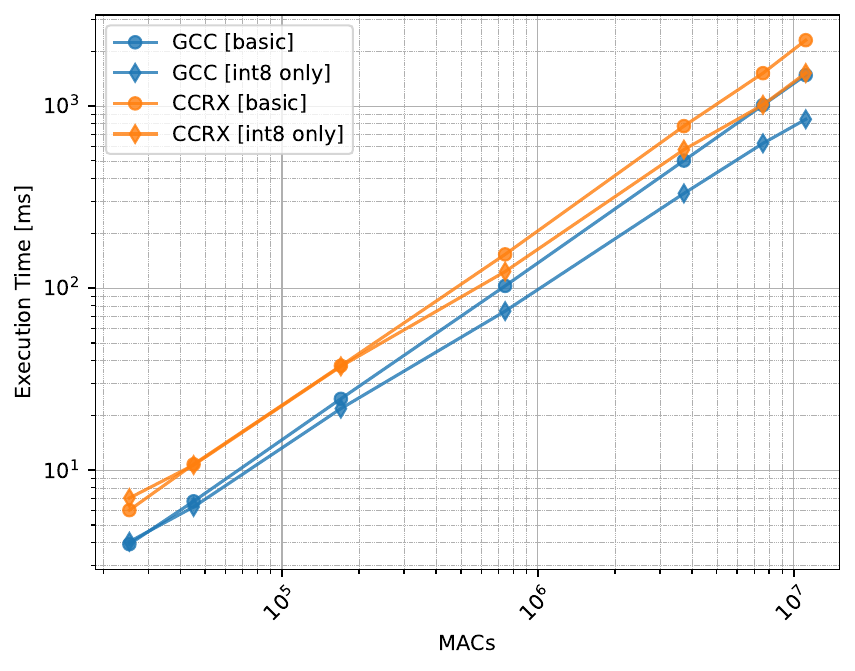}
    \caption{\Gls{cnn} models.}
    \label{fig:study_ccrx_cnn_exe}
  \end{subfigure}
  \caption{Execution time of models deployed using CC-RX and \gls{gcc} compilers on the Renesas RX65N board.}
  \label{fig:study_ccrx}
\end{figure*}

Renesas offers \gls{gcc} as a free compiler for their microcontrollers, while they also provide their proprietary compiler, CC-RX, which is further optimized for RX microcontrollers. In this study, we evaluated the performance of \gls{fc} and \gls{cnn} models deployed on the Renesas RX65N board using eAI Translator and CC-RX and \gls{gcc} compilers.

Since we could not interpret the memory requirements of the models using CC-RX, we only evaluated the execution time of the models. As evidenced by Fig.~\ref{fig:study_ccrx}, the \gls{gcc} compiler demonstrated better performance than CC-RX in terms of execution speed under default settings and highest optimization levels.

\bibliographystyle{elsarticle-num}
\bibliography{references}

\end{document}